\documentclass[lettersize,journal]{IEEEtran}
\usepackage{amsmath,amsfonts}
\usepackage{amsthm}
\usepackage{amssymb}
\usepackage[ruled,vlined,linesnumbered]{algorithm2e}
\usepackage{array}
\usepackage[caption=false,font=normalsize,labelfont=sf,textfont=sf]{subfig}
\usepackage{textcomp}
\usepackage{url}
\usepackage{verbatim}
\usepackage{graphicx}
\usepackage{cite}
\def\BibTeX{{\rm B\kern-.05em{\sc i\kern-.025em b}\kern-.08em
    T\kern-.1667em\lower.7ex\hbox{E}\kern-.125emX}}
\usepackage{multicol}
\usepackage[mathscr]{euscript}
\usepackage{wasysym}
\usepackage{multirow}
\usepackage[export]{adjustbox}
\usepackage{subcaption} 
\usepackage{url} 
\usepackage{diagbox}
\usepackage{booktabs}
\usepackage{tabularx}
\usepackage{comment}
\usepackage{balance}
\newtheorem{definition}{Definition}

\newcommand{\argmin}{\operatornamewithlimits{argmin}}
\def\argmin{\mathop{\rm argmin}}
\def\RR{\mathbb R}

\newcommand{\bzero}{\mathbf{0}}
\newcommand{\bA}{\mathbf{A}}

\newcommand{\bC}{\mathbf{C}}
\newcommand{\bD}{\mathbf{D}}
\newcommand{\bd}{\mathbf{d}}
\newcommand{\be}{\mathbf{e}}
\newcommand{\bE}{\mathbf{E}}

\newcommand{\bI}{\mathbf{I}}
\newcommand{\bL}{\mathbf{L}}
\newcommand{\bM}{\mathbf{M}}

\newcommand{\bO}{\mathbf{0}}

\newcommand{\bR}{\mathbf{R}}
\newcommand{\bS}{\mathbf{S}}

\newcommand{\bu}{\mathbf{u}}
\newcommand{\bU}{\mathbf{U}}
\newcommand{\bV}{\mathbf{V}}
\newcommand{\bv}{\mathbf{v}}

\newcommand{\bW}{\mathbf{W}}
\newcommand{\bx}{\mathbf{x}}
\newcommand{\bX}{\mathbf{X}}

\newcommand{\bY}{\mathbf{Y}}

\newcommand{\bbeta}{{\boldsymbol{\beta}}}
\newcommand{\bmu}{{\boldsymbol{\mu}}}
\newcommand{\bgamma}{{\boldsymbol{\gamma}}}
\newcommand{\bGamma}{{\boldsymbol{\Gamma}}}

\newcommand{\beps}{{\boldsymbol{\epsilon}}}

\newcommand{\bPsi}{{\boldsymbol{\Psi}}}

\newcommand{\btheta}{{\boldsymbol{\theta}}}

\newcommand{\bSigma}{{\boldsymbol{\Sigma}}}
\newcommand{\bpsi}{{\boldsymbol{\psi}}}
\newcommand{\bsig}{{\boldsymbol{sig}}}
\newcommand{\bSNR}{{\boldsymbol{SNR}}}
\newcommand{\bone}{{\boldsymbol{1}}}
\newcommand{\N}{{\cal N}}

\newcommand{\diag}{{\rm{diag}}}
\newcommand{\tr}{{\rm{tr}}}
\newcommand{\var}{{\rm{var}}}

\newcommand{\abs}[1]{\left|#1\right|}
\newcommand{\op}{\mathrm{op}}

\newtheorem{proposition}{Proposition}
\newtheorem{theorem}{Theorem}

\newtheorem{corollary}{Corollary}

\newtheorem{lemma}{Lemma}
\newtheorem*{lem}{Lemma}

\begin{document}
\title{Feature Selection for Latent Factor Models}
\author{Rittwika Kansabanik, Adrian Barbu}
\markboth{Transaction on Pattern Analysis and Machine Intelligence}
{Rittwika Kansabanik, Adrian Barbu,:
Feature Selection For Latent Factor Models}
\maketitle
\begin{abstract}
Feature selection is crucial for pinpointing relevant features in high-dimensional datasets, mitigating the 'curse of dimensionality,' and enhancing machine learning performance. 
Traditional feature selection methods for classification use data from all classes to select features for each class.
This paper explores feature selection methods that select features for each class separately, using class models based on low-rank generative methods and introducing a signal-to-noise ratio (SNR) feature selection criterion. 
This novel approach offers a theoretically sound guarantee of feature recovery under specific assumptions and has been demonstrated to outperform some existing feature selection methods on standard classification datasets. 
Additionally, it can be easily applied in a class-incremental learning setup, yielding performance comparable to that of some current methods. 
\end{abstract}



\section{Introduction}
\label{sec:intro}
Feature selection is a central component of modern data-driven systems in computer vision, signal processing, and related areas.
High-dimensional feature spaces offer rich representational capacity, but they also introduce redundancy, noise, and computational cost.
The goal of feature selection is to identify a subset of features that retains predictive accuracy while improving interpretability and efficiency.
Most standard approaches, whether filter, wrapper, or embedded, do this by optimizing a single objective that is evaluated over all classes at once \cite{dhal2022comprehensive}.
The subset obtained in this way is a compromise across classes: a feature is selected because it is useful for the overall decision rule, not because it characterizes any particular class.
Such models are also coupled across classes.
When a new class is added, the objective must be re-optimized, and the parameters that encoded the earlier classes are modified in the process.
This is one source of catastrophic forgetting in incremental settings.

This paper takes a different approach.
Each class is modeled independently by a low-rank generative model,
$\bx = \bmu + \bW \bgamma + \beps$,
and feature selection is performed separately for each class.
Given the parameters $(\bW, \bPsi)$ estimated for a class, its features are ranked by the signal-to-noise ratio
$\bSNR_i = (\bW\bW^T)_{ii} / \sigma^2_i$,
which compares the variance of feature $i$ that is explained by its latent factors to the variance that is not explained.
Features with high SNR are strongly associated with the latent structure of that class and are retained.
No data or parameters are shared across classes at any stage.

This design has a direct consequence for class-incremental learning.
Each class has its own model and its own selected features, and neither is updated once the class has been learned.
Prior knowledge is therefore preserved exactly rather than approximated, and learning a new class cannot degrade performance on earlier ones, because no parameters are shared between them.
The method requires no exemplars, no knowledge distillation, and no bias-correction stage, and the cost of adding a class does not grow with the number of classes already seen.

Low-rank and latent factor models have been used for feature selection in earlier work, for example in multi-label problems \cite{lim2023low} and in Alzheimer's disease diagnosis \cite{salas2010feature}, where factor loadings were used to identify relevant brain regions.
These studies confirm that latent factor models can separate signal from noise.
However, they are designed for static datasets, and they generally obtain sparsity either by imposing structural constraints on the model or by combining dimensionality reduction with a separate search procedure.
The SNR criterion requires neither.
It is computed directly from the estimated parameters, and the number of selected features can be changed by re-thresholding, without refitting the model.

Our earlier work \cite{kansabanik2025feature} introduced this criterion and showed that, for fixed $d$ and as $n \to \infty$, the estimated SNRs converge in probability to their true values and the correct features are recovered with probability tending to one.
The present paper extends that work in three directions.
First, it develops the class-incremental feature selection method implied by the per-class design and evaluates it against recent methods from the literature.
Second, it broadens the estimation machinery, adding a nonparametric latent factor estimator and extending scalable Mahalanobis scoring \cite{wang2023scalable} to models with heteroscedastic noise.
Third, it analyzes the estimators beyond the limiting regime, giving explicit low sample bounds for SNR convergence and consequently true feature recovery.

\noindent The main contributions of this paper are the following.
\begin{itemize}
\item A per-class SNR criterion for feature selection with low-rank generative models, instantiated with PPCA, LFA, HeteroPCA, and ELF, the latter being a nonparametric latent factor estimator introduced in this paper.

\item An exemplar-free class-incremental feature selection method that does not suffer from catastrophic forgetting and whose cost per added class does not grow with the number of classes.

\item A theoretical analysis of the SNR criterion. We characterize the stationary points of the LFA log-likelihood, derive the asymptotic order of the signal-variance and SNR estimation errors, and obtain a non-asymptotic bound on the signal-variance estimation error under the PPCA model.

\item An experimental evaluation on simulated data and on CIFAR-10, CIFAR-100, and ImageNet-1k using DINOv3 and CLIP features, with comparisons against FSA \cite{barbu2016feature} and TISP \cite{she2012iterative} and against recent class-incremental learning methods.
\end{itemize}
\section{Related Work}
\label{sec:RelatedWork}

\subsection{PCA \& LFA-Based and Hybrid Feature Selection}

While many techniques leverage Principal Component Analysis (PCA) for dimensionality reduction, these approaches to feature selection differ significantly from the proposed SNR-based method. 
Boutsidis et al.~\cite{boutsidis2008unsupervised} focused on selecting a representative subset of features that preserves the variance captured by top eigenfeatures. 
Our SNR-based method, however, evaluates each feature individually for its discriminatory power rather than selecting a collective subset to represent the whole. 
The work by Niu and Qiu~\cite{niu2010facial} on weighted PCA is conceptually extended by the SNR approach, which formalizes the weighting by using the inverse of the noise covariance to systematically distinguish meaningful signals from noise.

Several hybrid methods use PCA as a preliminary step before applying other selection techniques. 
For instance, \cite{uuguz2011two} employed a two-stage method combining Information Gain and a Genetic Algorithm (GA) with PCA. 
Similarly, Ahmad~\cite{ahmad2015feature}, Alomari et al.~\cite{alomari2022hybrid}, and Pushpalatha et al.~\cite{pushpalatha2023cardio} used evolutionary algorithms such as the GA, Grey Wolf Optimizer, and ReliefF to refine a PCA-reduced feature set. 
These multi-step approaches separate dimensionality reduction from feature selection, whereas our SNR-based method provides an integrated solution in which the selection criterion is inherent to the model.

Supervised PCA variants also differ in their core mechanism. 
Sharifzadeh et al.~\cite{sharifzadeh2017sparse} and Rahmat et al.~\cite{rahmat2024supervised} incorporate supervision by identifying features that are highly dependent on a response variable. 
In contrast, our SNR-based method models each class independently and selects features based on their ability to represent the unique characteristics of that class, not by their correlation with an output variable.

Feature selection using latent factor models has also seen diverse approaches. 
The Sparse Estimation of Latent Factors (SELF) framework proposed by Aziz~\cite{aziz2023sparse} achieves feature selection by imposing sparsity directly on the model's transformation matrix $\bW$. 
The SNR-based method diverges from this by removing such structural constraints and instead using the signal-to-noise ratio as a post-hoc criterion to rank features based on the learned model.

The work of Abbas and Sivaswamy~\cite{7486536} utilized latent factors to extract influential low-dimensional features from medical images, followed by classification using Mahalanobis distance. 
This constitutes a feature-extraction approach, creating new features rather than a feature-selection method that ranks and chooses from the original set of features, which is the focus of the SNR method. 
Similarly, Townes et al.~\cite{townes2019feature} proposed a latent factor model for single-cell RNA-sequencing data that ranks genes based on deviance, a metric tailored to count-based data. 
Our SNR-based method is more general, defining the signal and noise based on the variance explained by the generative model, making it applicable across various data types.

In conclusion, while PCA- and LFA-based methods often rely on hybrid frameworks or structural constraints for feature selection, the SNR-based approach offers a distinct, unified methodology. 
It evaluates features on a class-by-class basis using an intrinsic, theoretically grounded measure of their signal content.

\subsection{Signal-to-Noise Ratio (SNR)–Based Feature Selection}

The Signal-to-Noise Ratio (SNR) is one of the simplest and most interpretable measures for identifying discriminative features. 
It quantifies how strongly a feature separates classes relative to within-class variation, making it an effective filter criterion in high-dimensional data.

The use of SNR as a feature selection criterion dates back to early work in neural networks. 
\cite{bauer2000feature} introduced an SNR-based saliency measure to identify and prune noisy inputs during training by comparing each input's contribution to that of a random noise feature. 
This approach proved effective for dimensionality reduction on benchmark datasets, outperforming standard PCA-based methods. 
Shortly after, the now-classic SNR formulation, $SNR = (\mu_1 - \mu_2) / (\sigma_1 + \sigma_2)$, was employed by \cite{huang2003comparative} as a screening criterion for selecting features in binary classification tasks with probabilistic neural networks, where $(\mu_i, \sigma_i)$ represent the mean and standard deviation of a feature for class $i$.

The utility of SNR became particularly evident in bioinformatics, where microarray and gene expression data are characterized by high dimensionality and significant noise. 
To improve feature ranking, \cite{mishra2011feature} and later \cite{sahu2017feature} proposed hybrid clustering frameworks where SNR was used to select the most informative genes within each cluster, effectively reducing redundancy and enhancing classification accuracy.

In recent years, SNR-based methods have been hybridized with other machine learning techniques to enhance their robustness and applicability across diverse domains. 
In engineering, \cite{hamadache2017principal} used a PCA-based signal subspace approach to improve the SNR of noisy vibration signals for early fault detection in ball bearings. 
For wireless positioning systems, \cite{oh2024minimum} developed SNR-driven feature reduction techniques to minimize model complexity while maintaining predictive power in low-SNR environments.

In genomics, Weighted SNR (WSNR) methods have been developed, such as the one by \cite{hamraz2023feature}, which integrates SNR scores with Support Vector Machine (SVM) weights to emphasize the most discriminative genes. 
To mitigate the influence of outliers in skewed datasets, \cite{hussain2024optimal} combined SNR scores with the Mood median test, creating a robust "Md-score" that balances class separation and statistical significance. 
Beyond classification, SNR has been adapted for nonlinear regression in physical systems. 
\cite{bihl2025physics} introduced an ANN-SNR method with confidence interval stopping rules to predict concrete shear strength, achieving high accuracy with a significantly reduced feature set.

While existing SNR-based feature selection strategies are flexible, they often rely on heuristics and lack theoretical rigor. 
We address these limitations by defining SNR through the variance components of a generative latent factor model, implemented within a scalable, class-specific paradigm. 
Crucially, we establish a formal statistical foundation for this approach by deriving the asymptotic properties and non-asymptotic probability bounds of the SNR estimators.

\subsection{Feature Selection for Multi-class Classification}

Traditional techniques like Correlation-Based Selection \cite{khalid2014survey} and sparsity-driven pruning—including $l_0-$constraints \cite{guo2021network, barbu2016feature}, iterative thresholding \cite{she2009thresholding, she2012iterative}, and autoencoder-based reconstruction \cite{7727221}—often prioritize global subset relevance or sparsity over class-specific discriminatory signals. 
While supervised PCA \cite{yu2006supervised} and class-incremental PPCA \cite{wang2023scalable} incorporate label information, they remain limited by scalability issues or the absence of an explicit feature selection mechanism. 
Furthermore, hybrid transformation methods \cite{zhou2014face} lack interpretability and require full feature measurements because they rely on linear combinations of variables. 
In contrast, our proposed method employs a class-specific SNR criterion and Mahalanobis distance-based classification within distinct feature subspaces, achieving true parsimony while enhancing interpretability and operational efficiency.

\subsection{Asymptotic \& Non-asymptotic Analysis}

A central pursuit in modern high-dimensional statistics is to understand not only \textit{if} an estimator converges to its true value, but at what \textit{rate} it converges and with what level of certainty. 
Latent variable models, and in particular Latent Factor Analysis (LFA) and Probabilistic Principal Component Analysis (PPCA), represent a foundational toolkit for modeling and dimensionality reduction in complex datasets. 
For decades, the theoretical understanding of these models was primarily dominated by classical asymptotic results, which guarantee consistency and efficiency as the number of samples approaches infinity \cite{girshick1936principal,girshick1939sampling,anderson1963asymptotic,waternaux1976asymptotic}. 

However, while asymptotic results establish eventual consistency, they are insufficient for a rigorous analysis of an estimator's behavior in the non-limiting regime (finite $n$). 

In the context of our work, the reliability of any feature selection procedure based on ranking estimated SNR values fundamentally depends on a finite-sample bound. 
This bound establishes a high-probability "contract" that the estimated value $\hat{\bSNR}$ lies within a quantifiable neighborhood of the actual, unobservable $\bSNR$. 

Without such an explicit, non-asymptotic bound, the numbers produced by the estimator lack formal validation, making any subsequent analysis heuristic. 

This paper takes an initial step toward deriving such a bound. 
We move beyond the traditional asymptotic regime to analyze the explicit nature of the parameter estimation errors for moderate values of $n$. 
Our theoretical framework is built upon the powerful tools of modern probability theory, leveraging seminal results in matrix concentration inequalities and the behavior of quadratic forms of sub-Gaussian random vectors \cite{wainwright2019high,zhivotovskiy2024dimension}. 
The primary contribution of this work is to translate these abstract mathematical tools into concrete, interpretable guarantees for the key parameters of latent factor models. 
The implications are significant: these bounds provide a formal basis for model validation, deepen understanding of the statistical difficulty of parameter estimation in the presence of noise, and offer a principled way to reason about the reliability of any downstream task that depends on these models.

\subsection{Class-Incremental Learning}

Incremental learning (IL) is a paradigm designed to facilitate continuous adaptation to sequential data without the prohibitive computational costs of retraining on the entire dataset. 
This approach is essential for large-scale, dynamic environments, yet it is fundamentally challenged by catastrophic forgetting—the phenomenon where optimizing for new tasks overwrites parameter configurations essential for previous class recognition. 
This results in a precipitous accuracy drop for old classes, necessitating strategies that balance plasticity with stability.

The literature has bifurcated into two primary schools of thought: knowledge preservation through regularization, and the correction of inherent classifier biases. 
Knowledge distillation (KD), introduced by \cite{hinton2015distilling}, is a foundational technique for regularization that uses softened output probabilities to stabilize decision boundaries. 
Building upon this, \cite{zagoruyko2016paying} utilized activation maps to transfer knowledge between model iterations. 
The iCaRL framework \cite{rebuffi2017icarl} integrated these concepts by combining dynamic exemplar management with a nearest-class-mean classifier to mitigate forgetting. 
Further refinements include the attention distillation loss proposed by \cite{dhar2019learning} for data-free knowledge transfer and the PODNet architecture \cite{douillard2020podnet}, which employs multi-scale feature distillation across spatial resolutions to preserve semantic information.

Parallel to representation preservation, other researchers focus on mitigating the recency bias toward newly introduced classes. 
For instance, \cite{hou2019learning} introduced UCIR, which replaces the standard softmax with cosine normalization to address class imbalance. 
Similarly, BiC \cite{wu2019large} employs a two-stage training approach to estimate and adjust task bias within the model’s final layers.

Recent advancements increasingly leverage large-scale Pre-trained Models (PTMs). 
\cite{nakata2022revisiting} utilized the CLIP \cite{radford2021clip} architecture to implement a robust KNN-based classifier. 
Following a similar trajectory, \cite{wang2023scalable} demonstrated that frozen self-supervised features combined with class-specific generative PPCA models can structurally eliminate interference between tasks. 
This emerging domain continues to expand through methods like prototype classification \cite{zhou2025revisiting} and knowledge rumination \cite{Gao_2025_CVPR}, which aim to maximize generalization and adaptability while minimizing catastrophic forgetting in class-incremental settings.

\section{Signal to Noise Ratio (SNR) for Feature Selection} 
\label{sec:SNR}

This section introduces a feature selection technique that uses SNR as the feature selection criterion. 
This method can be used for various low-rank generative models such as Probabilistic PCA and Latent Factor Analysis. 
First, we describe these methods and their parameter estimation processes. 
We then use these estimates to calculate the SNRs.

\vspace{-1mm}
\subsection{Notations}
\vspace{-1mm}

We denote matrices by uppercase bold letters such as $\bM\in \RR^{p \times q}$, vectors with lowercase bold letters such as $\bv \in \RR^d$ and scalars by lowercase letters, e.g. $x\in \RR$. 
$\bM_{ij},\bM_{\cdot j}\text{ and }\bM_{i\cdot}$ represent the $(i,j)^{th}$ element, $j^{th}$ column and $i^{th}$ row of $\bM$ respectively. 
$\bv_i \text{ and }\bv_{(i)}$ denote $i^{th}$ element and the $i^{th}$ largest value of vector $\bv$. 
$m$ is used to specify the number of selected features. 
$d \text{ and } n$ denote the number of available features and the number of observations, respectively, for a given class. 
$\diag(a_1,a_2,\cdots, a_d)$ denotes a $d \times d$ diagonal matrix with diagonal elements $\{a_1,a_2, \cdots, a_d\}$. 
$D(\bM)$ denotes a diagonal matrix with the same diagonal elements as $\bM$. 
$\bSNR$ denotes a vector containing the SNR values of all available features. 

\vspace{-1mm}
\subsection{Low-rank Generative Models}
\label{sec:lrmodels}
\vspace{-1mm}

In this section, we describe the four different methods based on low-rank generative models that are included in this study, namely Probabilistic PCA (PPCA) \cite{tipping1999probabilistic}, Latent Factor Analysis (LFA)\cite{ghahramani1996algorithm}, Heteroskedastic PCA (HeteroPCA) \cite{zhang2022heteroskedastic} and Estimation of Latent Factors (ELF). 
The last method is a novel nonparametric version of LFA, and is first introduced in this paper. 

PPCA, LFA, and our newly introduced method, ELF, share the same model structure but differ in the assumptions made about their model parameters. 
The model aims to find a relationship between the observed $\bx\in \RR^d$  and a hidden set of variables (latent variables) $\bgamma\in \RR^r$ with $r << d$, and assumes the latent factors and noise variables are independent of each other. 
It is as follows:
\vspace{-2mm}
\begin{equation} \label{eq:LFAm}
\bx = \bmu + \bW \bgamma + \beps, \quad \text{with $E(\beps) = \bzero$ and } \var(\beps) = \bPsi.
\vspace{-2mm}
\end{equation}

PPCA and LFA methods assume that $\bgamma \overset{i.i.d}{\sim} \N(\bzero, \bI_r)$ and that the noise variable $\beps\overset{i.i.d}{\sim} \N(\bzero,\bPsi)$ . 
It can be easily verified that for these two methods:
\vspace{-3mm}
\begin{equation}\label{eq:distLFA}
\bx\vert \bgamma \sim \N(\bW\bgamma + \bmu , \bPsi), \text{ and by integration,}
\vspace{-2.5mm}
\end{equation}
\begin{equation}
    \bx \sim \N(\bmu, \bSigma), \bSigma = \bW\bW^T + \bPsi.
\vspace{-2mm}
\end{equation}

Conversely, ELF does not make distributional assumptions about the parameters it estimates. 
ELF assumes $\bGamma = (\bgamma_1,\bgamma_2,\cdots,\bgamma_n)^T$ to be semi-orthogonal ($\bGamma^T\bGamma = \bI_r$). 
LFA and ELF, while sharing similar goals with PPCA, assume distinct noise variances across dimensions.

\begin{equation}\label{eq:psi-def}
    \bPsi = 
    \begin{cases}
         \sigma^2 \bI_d & \text{for PPCA,}\\
          \diag(\sigma^2_1, \sigma^2_2, \cdots, \sigma^2_d) & \text{otherwise.}\\
    \end{cases} 
\end{equation}
$\bmu$ has been treated as a constant vector in the model \eqref{eq:LFAm} and is estimated as: $\bmu_{ML} = \frac{1}{n}\sum_{i=1}^n \bx_i$.

\noindent{\bf Parameter estimation for PPCA.}
A closed form of the Maximum Likelihood (ML) estimates of the PPCA model parameters $(\bW, \sigma^2)$ was given in \cite{tipping1999probabilistic}, as follows:

\begin{equation}
\label{eq:cest}
\begin{split}
    \sigma^2_{ML} &= \frac{1}{d-r}\sum_{j = r+1}^d \lambda_j, \\ 
    \bW_{ML} &= \bU_r(\bS_r - \sigma^2_{ML}\bI_r)^{0.5} \bR,
\end{split}
\end{equation}

where $\lambda_j$ is the $j^{th}$ largest eigenvalue and $\bU_r$ consists of the first $r$ principal eigenvectors of the sample covariance matrix $\hat{\bSigma}$. 
The matrix $\bS_r = \diag(\lambda_1,\lambda_2,\cdots,\lambda_r)$, while $\bR$ is an arbitrary $r \times r$ orthogonal rotation matrix. 

\noindent{\bf Parameter estimation for LFA.}
LFA parameters $(\bW, \bPsi)$ can be estimated using an EM algorithm due to \cite{ghahramani1996algorithm}.

\begin{theorem}[due to \cite{ghahramani1996algorithm}]\label{thm:lfa}
Assume that the data has been properly centralized and let $\bbeta = \bW^T(\bPsi + \bW\bW^T)^{-1}$. 
The EM updates of $(\hat{\bW}, \hat{\bPsi})$ for LFA are:
\begin{itemize}
\item \textbf{E-step}: Compute $E(\bgamma|\bx_i)$  and $E(\bgamma\bgamma^T|\bx_i)$ for each data point $\bx_i$ as follows:
\[
E(\bgamma|\bx_i) = \bbeta\bx_i, \; E(\bgamma\bgamma^T|\bx_i) = \bI_r - \bbeta \bW + \bbeta\bx_i\bx_i^T\bbeta^T.
\]
\item \textbf{M-step} Update the LFA parameters as:
\vspace{-2mm}
\begin{equation}\label{eq:LFA-EM}
\bW_{new} = \sum_{i=1}^n \bx_iE(\bgamma|\bx_i)^T(\sum_{i=1}^n E(\bgamma\bgamma^T|\bx_i))^{-1},   
\vspace{-4mm}
\end{equation}
\begin{equation}
\bPsi_{new} = \frac{1}{n}D(\sum_{i=1}^n\bx_i\bx_i^T- \bW_{new}E(\bgamma|\bx_i)\bx_i^T).
\vspace{-2mm}
\end{equation}
\end{itemize}
\end{theorem}

\noindent{\bf Parameter estimation for ELF.}
ELF estimates the model parameters $(\bGamma,\bW)$ by optimizing the following:
\vspace{-2mm}
\begin{equation} \label{ELF}
(\hat{\bW}_{ELF}, \hat{\bGamma}_{ELF})
= \hspace{-3mm} \argmin_{(\bGamma, \bW),\bGamma^T \bGamma = \bI_r} \hspace{-3mm}\|(\bX - \bGamma \bW^T)\bPsi^{-\frac{1}{2}} \|^2_F,
\vspace{-2mm}
\end{equation}
where $\bX = (\bx_1-\bmu_{ML},\bx_2-\bmu_{ML},\cdots,\bx_n-\bmu_{ML})^T$, i.e. properly centralized, $\bPsi=\diag(\hat{\sigma}^2_1,...,\hat{\sigma}^2_d)$, and  $\hat{\sigma}^2_j = \|\hat{\bX}_{\cdot j} -\bX_{\cdot j}\|_2^2/(n-1)$.
We use  $\bPsi^{-\frac{1}{2}}$ as feature weights in \eqref{ELF} to reduce the impact of features with significant unexplained noise variance, thereby significantly improving model accuracy. 
To perform the minimization in \eqref{ELF}, in every iteration, we first estimate $(\hat{\bW},\hat{\bGamma})$ without the constraint on $\bGamma$ using Theorem \ref{Thm:2} and then adjust the estimated parameters to satisfy the constraint using the Proposition \ref{prop:ELF} below. 
The proofs are included in the Supplementary Material.
\begin{theorem}\label{Thm:2}
The ELF objective \eqref{ELF} without the constraint $\bGamma^T \bGamma = \bI_r$ is minimized w.r.t $\bGamma$ and  $\bW$ by
\begin{equation}\label{elf est}
\hspace{-2mm}\hat{\bGamma} \hspace{-1mm}= \hspace{-1mm}\bX \bPsi^{-1}\hspace{-0.5mm}\bW\hspace{-0.5mm}(\hspace{-0.5mm}\bW^T \hspace{-0.5mm}\bPsi^{-1}\hspace{-0.5mm} \bW)^{-1} \text{and } 
\hat{\bW} \hspace{-1mm}= \hspace{-1mm}\bX^T\bGamma(\bGamma^T \bGamma)^{-1}\hspace{-2mm}. 
\end{equation}
\end{theorem}
\begin{proposition}\label{prop:ELF}
If \( \mathbf{U} \mathbf{D} \mathbf{V}^T = \mathbf{\Gamma} \) is the SVD of \(\mathbf{\Gamma}\), then \(\mathbf{\Gamma}_1 = \mathbf{U}\), \text { and } \(\mathbf{W}_1 = \mathbf{W} \mathbf{V} \mathbf{D}\) satisfy \(\mathbf{\Gamma}_1 \mathbf{W}_1^T = \mathbf{\Gamma} \mathbf{W}^T\) along with \(\mathbf{\Gamma}_1^T \mathbf{\Gamma}_1 = \mathbf{I}_r\).
\end{proposition}
The $\hat{\bW}$  produced in \eqref{elf est} does not depend on the feature weights $\bPsi$. 
Algorithm \ref{algo:ELF} summarizes the iterative estimation procedure.

\begin{algorithm}[ht!]
\caption{Parameter Estimation for ELF}
\label{algo:ELF}

\KwIn{data matrix $\bX \in \mathbb{R}^{n \times d}$, max iterations $T$, dimension $m$}
\KwOut{Estimated parameters $\hat{\bW}$ and $\hat{\bPsi}$}

\textbf{Initialization:} \\
\Indp
$\bPsi \gets \boldsymbol{I}_d$\tcp*{feature weight matrix} 
$\bGamma \gets$ top-$r$ principal components of $\bX$\;
$\bW \gets$ top-$r$ PCA loading vectors from $\bX$\;
\Indm

\For{$t = 1$ \KwTo $T$}{
    $\bW \gets \bX^\top \bGamma (\bGamma^\top \bGamma)^{-1}$ \tcp*{Update}
    
    $\bGamma \gets \bX \bPsi^{-1} \bW (\bW^\top \bPsi^{-1} \bW)^{-1}$ \tcp*{Update}
    
    Perform SVD: $\bGamma = \bU_{\bGamma} \bD_{\bGamma} \bV_{\bGamma}^\top$\;
    Update $\bGamma \gets \bU_{\bGamma}$, \quad $\bW \gets \bW \bV_{\bGamma} \bD_{\bGamma}$

    \For{$i = 1$ \KwTo $d$}{
        $\hat \sigma_i^2 \gets \mathrm{Var}(\bX_{\cdot i} - \bGamma \bW^\top_{i\cdot})$
    }
    $\bPsi \gets \mathrm{diag}(\hat \sigma_1^2, \ldots, \hat \sigma_d^2)$ \tcp*{Update}

    \If{$\|\bX - \bGamma \bW^\top\|_F$ is sufficiently small}{
        \textbf{break}
    }
}

\end{algorithm}

\noindent{\bf HeteroPCA.} HeteroPCA \cite{zhang2022heteroskedastic} addresses the issue of performing PCA when the data $\bX_{n \times d}$ has heteroskedastic noise in a spiked covariance model setup.  
It assumes the following setup:
\vspace{-3mm}
\begin{equation}\label{model:heteroPCA}
\bX_{n\times d} = \bX_0 + \beps, E(\bX_0) = \bmu, 
 Cov(\bX_0) = \bSigma_0, 
\vspace{-2mm}
\end{equation}
\begin{equation}
E(\beps) = 0, Cov(\beps) = \bPsi = \diag(\sigma^2_1, \sigma^2_2, \cdots, \sigma^2_d).
\vspace{-1mm}
\end{equation}
Here, $\bX_0$ is the noise-free version (signal) of the given data matrix $\bX$, and $\beps$ and $\bX_0$ are independent. 
$\bSigma_0$ admits a rank-$r$ ($<<d$) eigen-decomposition, i.e. $\bSigma_0 = \bU \bD \bU^T$ with $\bU \in \RR^{d \times r}$ and $\bD \in \RR^{r \times r}$. 
The goal is to estimate $\bU$.

Though the model is similar to LFA, the objective of Hetero PCA aims to capture the principal components (PCs) (known as $\hat{\bU}$) of the signal, accounting for heteroskedasticity. 
Since $E(\hat{\bSigma}) = \bSigma_0 + \bPsi$, there will be a significant difference between the principal components of $E(\hat{\bSigma})$ and those of $\bSigma_0$. 
To cope with the bias on the diagonal of $E(\hat{\bSigma})$, HeteroPCA iteratively updates the diagonal entries based on the off-diagonals so that the bias is significantly reduced and a more accurate estimation is achieved.
\vspace{-5mm}
\subsection{Estimation of SNR}
\vspace{-1mm}
The estimated signal-to-noise ratio (SNR) for the available features is computed based on the estimated $(\hat{\bW},\hat{\bPsi})$ obtained by the different methods. 
The SNR for the $i$-th feature is defined as:
\vspace{-2mm}
\begin{equation}\label{def:SNR}
\bSNR_i = \frac{\sum_{j=1}^r \bW_{ij}^2}{\sigma^2_i}, i \in \{1,2,\cdots, d\}.
\vspace{-1mm}
\end{equation}
Though the SNR can be directly calculated using Eq. \eqref{def:SNR} for PPCA, LFA, and ELF methods, the same is not valid for HeteroPCA. 
For HeteroPCA, we obtain the $r$ principal loading vectors $\hat{\bU}$, corresponding to $\bX_0$. 
To evaluate \eqref{def:SNR} for HeteroPCA, the following steps are followed to obtain $(\hat{\bW},\hat{\sigma}^2_1,\cdots,\hat{\sigma}^2_d)$:
\begin{itemize}
    \item \textbf{Estimation of signal strength:} The initial estimates are:  $\Tilde{\bGamma} = \bX\hat{\bU}$ and $\Tilde{\bW} = \hat{\bU}$. Next, Proposition \ref{prop:ELF} is used to obtain semi-orthogonal $\hat{\bGamma}$ and the corresponding $\hat{\bW}$. 
    
    \item \textbf{Estimation of noise variance:} Next, the estimatated $(\sigma^2_i, i = 1,2,\cdots,d)$ are obtained, using $\hat{\bX}_0 = \hat{\bGamma}\hat{\bW}^T$:
\vspace{-3mm}
\begin{equation}
\hat{\sigma^2_i} = \|(\hat{\bX}_0)_{i\cdot} -\bX_{i\cdot }\|_2^2/(n-1).    
\end{equation}
\end{itemize}
\vspace{-2mm}
The intuition for employing SNRs to identify key features in the latent factor model is based on the assumption that the data originates from a lower-dimensional latent space. 
The signal is represented as $\bW\bgamma$. 
The variance of the corresponding signals is captured by the diagonal elements of $\bW\bW^T$ or the row sum of squares of $\bW$. 
At the same time, the unexplained noise variance is reflected in the diagonal elements of $\bPsi$. 
Once the SNRs are estimated, feature selection is performed by employing a simple thresholding technique, as described in Algorithm \ref{algo:snr}.

\begin{algorithm}[ht!]
\caption{SNR-Based Feature Selection}
\label{algo:snr}

\KwIn{estimated loadings $\hat{\bW} \in \mathbb{R}^{d \times r}$, noise variances $\hat{\sigma}^2_1, \ldots, \hat{\sigma}^2_d$, number of selected features $m$}
\KwOut{$I_m$ — indices of the top $m$ selected features}

\For{$i = 1$ \KwTo $d$}{
    Compute $\mathrm{SNR}_i \gets \dfrac{\sum_{j=1}^{r} \hat{W}_{ij}^2}{\hat{\sigma}^2_i}$
}

Sort SNR values: $\mathrm{SNR}_{(1)} \leq \mathrm{SNR}_{(2)} \leq \cdots \leq \mathrm{SNR}_{(d)}$\;

Select top-$m$ features: $I_m \gets \left\{ i \in \{1,\ldots,d\} \mid \mathrm{SNR}_i \geq \mathrm{SNR}_{(d - m + 1)} \right\}$

\end{algorithm}

\section{Theoretical Analysis}
\label{sec:theory}

In our work, the reliability of any feature selection procedure based on ranking estimated SNR values fundamentally depends on the fidelity of the SNR estimates themselves.  A finite-sample bound establishes a high-probability "contract" that the estimated value $\hat{\bSNR}$ lies within a quantifiable neighborhood of the actual, unobservable $\bSNR$.

Our theoretical framework is built upon the powerful tools of modern probability theory, leveraging seminal results in matrix concentration inequalities and the behavior of quadratic forms of sub-Gaussian random vectors \cite{wainwright2019high,zhivotovskiy2024dimension}. 

\begin{theorem}[\textbf{Refined Bound for Sample Variance Error}]
\label{thm:scalar_bernstein}
Let $x_{1i},\dots,x_{ni}$ be i.i.d.\ with $x_{ki} \sim \N(0,\bSigma_{ii})$, and let
$\hat{\bSigma}_{ii} = \frac{1}{n}\sum_{k=1}^{n} x_{ki}^2$.
Let $c$ be the universal constant from Bernstein's inequality. 
For any failure probability $\delta \in (0, 1)$, if the sample size $n$ satisfies
\vspace{-2mm}
\begin{equation}
    n > \frac{8}{c} \ln(4/\delta),
\vspace{-2mm}
\end{equation}
then the sample variance error is bounded with probability at least $1-\delta/2$ by:
\vspace{-2mm}
\begin{equation}
    \left|\hat{\boldsymbol{\Sigma}}_{ii} - \boldsymbol{\Sigma}_{ii}\right| \le \sqrt{\frac{2}{c}} \bSigma_{ii} \cdot \sqrt{\frac{\ln(4/\delta)}{n}},
\vspace{-1mm}
\end{equation}
\end{theorem}
This theorem ensures that the error in estimating the variance of a quiet, low-variance feature will not be artificially inflated by the presence of a noisy, high-variance feature elsewhere in the data. 
This local control will be the key to analyzing the parameter estimates of LFA.

\subsection{PPCA}
\label{sec:ppca}

We first analyze the PPCA model.
Let $\bx_1,\dots,\bx_n$ be $n$ i.i.d.\ $d$-dimensional samples from the PPCA model
$\bx_k = \bW\bgamma_k + \beps_k$, under the following assumptions.
\begin{itemize}
\item[(\textbf{A1})] $\bgamma_k \sim \N(\bzero,\bI_r)$ and $\beps_k \sim \N(\bzero,\sigma^2\bI_d)$.
\item[(\textbf{A2})] The vectors $\bgamma_k$ and $\beps_{k'}$ are independent for all
$k,k' \in \{1,\dots,n\}$.
\end{itemize}

Under \textbf{(A1)}--\textbf{(A2)} the observations satisfy $\bx_k \sim \N(\bzero,\bSigma)$ with
$\bSigma = \bW\bW^\top + \sigma^2\bI_d$.
The isotropic noise assumption also yields the closed-form estimates \eqref{eq:cest}, which are
based on the eigendecomposition of the sample covariance matrix
$\hat{\bSigma} = \sum_{j=1}^{d} l_j \bu_j \bu_j^\top$, with $l_1 \ge \dots \ge l_d$.
We bound the error in the estimated signal variance of each feature, $(\widehat{\bW\bW^\top})_{ii}$, which are the quantities the SNR criterion depends on.

\begin{theorem}[\textbf{PPCA Signal Variance Estimation Error Bound}]\label{thm:ppca-signal}
Suppose \textbf{(A1)}--\textbf{(A2)} hold. Let $\delta \in (0,1)$ and let the sample size satisfy
\begin{equation}
    n > \max\left\{\frac{8}{c}\ln(4/\delta),\; 4r(\bSigma) + \ln(2/\delta)\right\}.
\label{eq:ppca-n-cond}
\end{equation}
Then, for a given $i \in \{1,\dots,d\}$, with probability at least $1-\delta$,
\begin{equation}
\begin{split}
    \big| (\bW\bW^\top)_{ii} - (\widehat{\bW\bW^\top})_{ii} \big|
    &\le \bSigma_{ii}\sqrt{\frac{2\ln(4/\delta)}{cn}} \\
    &\quad + \min\{\epsilon(\delta/2),\, 2\bSigma_{ii}\},
\end{split}
\end{equation}
where
\begin{equation}
\epsilon(\delta) = C_0\,\|\bSigma\|_{\op}
\sqrt{\frac{4r(\bSigma) + \ln(1/\delta)}{n}},
\qquad
r(\bSigma) = \frac{\tr(\bSigma)}{\|\bSigma\|_{\op}},
\end{equation}
and $C_0 = 160/3$.
\end{theorem}

\noindent
The bound has two terms of different character.
The first is the local sample-variance error of Theorem \ref{thm:scalar_bernstein}, applied at
level $\delta/2$; it scales with the variance $\bSigma_{ii}$ of the feature under consideration.
The second controls the error in the estimated noise level $\hat{\sigma}^2$ together with the
residual term arising from the discarded eigendirections, both of which are governed by the
operator-norm deviation $\|\hat{\bSigma}-\bSigma\|_{\op}$ and are therefore common to all features.
The constant $C_0 = 160/3$ comes from applying Theorem \ref{thm:zhivotovskiy} with
$\kappa^2 = 8/3$, which is the value attained by centered Gaussian vectors.

\noindent\textbf{Dependence on the dimension.}
The bound involves $d$ only through the effective rank
$r(\bSigma) = \tr(\bSigma)/\|\bSigma\|_{\op}$, a consequence of the dimension-free form of
Theorem \ref{thm:zhivotovskiy}.
When the spectrum of $\bSigma$ is concentrated in a few directions, $r(\bSigma)$ is small and
condition \eqref{eq:ppca-n-cond} can be met with $n \ll d$.
The signal variance can then be estimated accurately in the regime $d \gg n$, provided the data
are approximately low-rank.

\noindent\textbf{Two regimes.}
The second term equals $2\bSigma_{ii}$ when
\begin{equation}
n < \frac{C_0^2\|\bSigma\|_{\op}^2}{4\bSigma_{ii}^2}\big(4r(\bSigma)+\ln(2/\delta)\big)
=: n_i(\delta),
\label{eq:ppca-threshold}
\end{equation}
and equals $\epsilon(\delta/2)$ otherwise.
Below $n_i(\delta)$ the bound is of the same order as the signal itself and is therefore
uninformative.
Above it the bound is governed by $\epsilon(\delta/2)$, decays at the parametric rate
$O(1/\sqrt{n})$, and vanishes as $n \to \infty$.
The quantity $n_i(\delta)$ may be read as a reliability threshold for feature $i$.

\noindent\textbf{Implications for feature selection.}
The threshold \eqref{eq:ppca-threshold} scales as $\|\bSigma\|_{\op}^2/\bSigma_{ii}^2$, so it is
smallest for features whose variance is comparable to the largest eigenvalue of $\bSigma$ and
largest for weak features.
The bound is thus informative at moderate sample sizes precisely for the high-variance features,
and remains uninformative for the low-variance ones, which are the candidates for removal.
We note that $\epsilon(\delta/2)$ does not depend on $i$: the guarantee is on the error relative
to $\bSigma_{ii}$ rather than on a smaller absolute error for strong features.
The result therefore supports the use of $(\widehat{\bW\bW^\top})_{ii}$ as a selection criterion
in the sense that the estimates of the leading features are the first to become reliable as $n$
grows; it does not by itself certify the ordering of the features, which we address next.

\noindent\textbf{From estimation to recovery.}
Theorem \ref{thm:ppca-signal} bounds the error of each estimated signal variance, but a selection
procedure is correct only if the estimated values order the features correctly across the boundary
of the selected set.
This requires the true values to be separated by more than the estimation error, as follows.

\begin{corollary}[\textbf{Non-Asymptotic Feature Recovery for PPCA}]
\label{cor:ppca-recovery}
Suppose the following hold for some $\delta \in (0,1)$.
\begin{itemize}
\item The PPCA model assumptions (\textbf{A1})--(\textbf{A2}).
\item There are $m$ true features with indices
$S = \{s_1, s_2, \dots, s_m\}$ and $(d-m)$ noisy features, i.e.
\begin{equation}
\theta_{s} > 0 \;\;\text{for } s \in S,
\qquad
\theta_{j} = 0 \;\;\text{for } j \notin S.
\label{eq:ppca-sparsity}
\end{equation}
Here $\theta_i = (\bW\bW^\top)_{ii}$ and $\hat{\theta}_i = (\widehat{\bW\bW^\top})_{ii}$, denote the
true and estimated signal variances of feature $i$.
\item The sample size satisfies
\begin{equation}
n > \max\left\{\frac{8}{c}\ln(4d/\delta),\; 4r(\bSigma) + \ln(2/\delta)\right\}.
\label{eq:ppca-rec-n-cond}
\end{equation}
\item The separation condition
\begin{equation}
\min_{s \in S} \theta_{s} - \max_{j \notin S} \theta_{j}
= \min_{1 \le k \le m} \theta_{s_k}
\;>\; 2B(\delta).
\label{eq:ppca-gap}
\end{equation}
where 
\begin{equation}
B(\delta) = \bSigma_{\max}\sqrt{\frac{2\ln(4d/\delta)}{cn}}
+ \min\{\epsilon(\delta/2),\, 2\bSigma_{\max}\}.
\label{eq:ppca-Bmax}
\end{equation}
\end{itemize}
Then, with probability at least $1-\delta$,
\begin{equation}
\big\{ i : \hat{\theta}_{(i)} \ge \hat{\theta}_{(d-m+1)} \big\} = S = \{s_1, s_2, \dots, s_m\},
\label{eq:ppca-recovery-claim}
\end{equation}
that is, the $m$ largest values of $(\widehat{\bW\bW^\top})_{ii}$ are attained exactly at the
true features $s_1, s_2, \dots, s_m$.
Here, $\hat{\theta}_{(k)}$ denotes the $k$-th
smallest value of $\{\hat{\theta}_1,\dots,\hat{\theta}_d\}$; $\bSigma_{\max} = \max_{i}\bSigma_{ii}$;
$\epsilon(\cdot)$, $r(\bSigma)$ and $C_0$ are as in Theorem \ref{thm:ppca-signal}; $c$ is the
constant of Theorem \ref{thm:scalar_bernstein}.
The same conclusion holds for selection by the estimated SNR, in which case \eqref{eq:ppca-gap}
reads
\begin{equation}
\min_{1 \le k \le m} \bSNR_{s_k} \;>\; \frac{2B(\delta)}{\sigma^2}.
\label{eq:ppca-gap-snr}
\end{equation}

\end{corollary}

\noindent\textbf{On the sample size condition.}
The $\ln d$ in \eqref{eq:ppca-rec-n-cond}, absent from \eqref{eq:ppca-n-cond}, is the price of the
union bound: feature recovery requires controlling the error of all $d$ features simultaneously, not of a
single fixed one.
Since $B(\delta)$ grows as $\sqrt{\ln d}$ and decays as $1/\sqrt{n}$, the dependence is mild, but
\eqref{eq:ppca-gap} becomes harder to satisfy as $d$ grows with $n$ fixed, consistent with the
decrease in selection accuracy with $d_{noise}$ reported in Table \ref{tab:Sim1}.

\noindent
Condition \eqref{eq:ppca-gap} is the finite-sample counterpart of the separation assumption used in
\cite{kansabanik2025feature}, where the gap need only be positive because the estimation error
vanishes as $n \to \infty$.
Since $B(\delta) = O(\sqrt{\ln(d)/n})$ once $n$ exceeds \eqref{eq:ppca-threshold}, any fixed
positive gap satisfies \eqref{eq:ppca-gap} for $n$ large enough, recovering the asymptotic
statement.
\subsection{LFA}
We now turn to the more general Latent Factor Analysis (LFA) model, where the idiosyncratic noise $\bPsi$ is diagonal but not necessarily isotropic. 
This flexibility prevents a simple closed-form solution for the ML estimators, which are instead found using Expectation-Maximization (EM), as described in Theorem \ref{thm:lfa}. 
Our goal is to analyze the behavior of the estimation error for feature-specific parameters: the signal variances $(\bW\bW^T)_{ii}$ and the noise variances $\bPsi_{ii}$.

One of the foundational asymptotic theories for the high-dimensional LFA problem was comprehensively established in the past by \cite{bai2012statistical}. 
Their main contribution is the development of a complete asymptotic theory for the MLE of LFA parameters when $(n,d) \to \infty$. 
To achieve this goal, they have considered some assumptions on the model parameters:
\begin{itemize}
\item[\textbf{(A3)}] The latent factors $\{\bgamma_k\}$ are either fixed constants
or random and independent of $\{\beps_k\}$, with sample covariance
$M_{\Gamma\Gamma} = \frac{1}{n}\sum_{k=1}^{n}\bgamma_k\bgamma_k^\top$ converging to a positive
definite limit.
In the random case the analysis of \cite{bai2012statistical} is understood as conditional on
$\{\bgamma_k\}$, as noted in Section 2 of that paper.

    
\item[\textbf{(A4)}] The noise $\{\beps_i,i = 1,2,\dots,n\}$ are iid random variables, and they are independent also with  $\{\bgamma_i, i = 1,2,\dots,n\}$. Also, $E(\beps_i) = \bzero$, $E(\beps_i\beps_i^\top) = \bPsi$, and $E(\epsilon_{ij}^4)\le C^4$ for all $j$ where $\bPsi = \diag(\sigma_1^2, \sigma_2^2, \dots \sigma_d^2)$.
    
\item[\textbf{(A5)}] For some positive constant $C$:
\begin{itemize}
    \item $\|\bW_{j\cdot}\|_2 \le C $,
    \item $\bW^T \bPsi^{-1}\bW/d \to \bC$,
    \item $\lim_{d \to \infty} \frac{1}{d} \sum_{j=1}^{d}\sigma_j^{-4}[\bW_{j\cdot} \otimes \bW_{j\cdot}][\bW_{j\cdot}^T \otimes \bW_{j\cdot}^T] = \Omega$,
\end{itemize}
where $(\bC,\Omega)$ are positive definite matrices. 
This ensures that influential, detectable factors are present in high dimensions. 
    
\item[\textbf{(A6)}] The estimated variances $\hat\sigma_j^2$ are constrained to the compact set $[\sigma_{\min}^2, \sigma_{\max}^2]$, a standard technical requirement for the consistency of nonlinear estimators.
\end{itemize}
The results of \cite{bai2012statistical} provide a clear idea of the convergence rates for $\hat{\bW}$ and $\hat{\bPsi}$. 

Assumption \textbf{(A5)} controls how the signal is distributed across the features. Since $\bW^\top\bPsi^{-1}\bW = \sum_{j=1}^{d}\sigma_j^{-2}\bW_{j\cdot}\bW_{j\cdot}^\top$ is a sum of
per-feature contributions, the positive definiteness of $\bC$ in \textbf{(A5)} requires the noise-weighted signal to
accumulate at rate $d$, so that each factor is present in a non-vanishing fraction of the features.
Positive definiteness of $\Omega$ requires the same of every pair of factors.
Together the two conditions state that the factors are \emph{pervasive}, which is what makes them
detectable as the number of features grows.

Pervasiveness is what makes a growing number of features an advantage rather than a difficulty.
Each feature that loads on a factor provides an independent, noisy view of it, so a factor observed
through many features can be estimated more accurately than one observed through few.
This is why the limit $d \to \infty$ is informative in the theory of \cite{bai2012statistical}:
the number of observations of each factor grows with $d$.
If instead the signal is confined to a fixed number of features, the additional features contribute
noise but no information about the factors, the sums in \textbf{(A5)} remain bounded while
being divided by $d$, and the limits degenerate.

The condition therefore stands in a certain tension with the premise of feature selection.
Concretely, if the number of relevant features is fixed at $m$ while $d \to \infty$, so that
$\bW_{j\cdot} = \bzero$ for all but the $m$ features indexed by $S$, then by \textbf{(A5)} and
\textbf{(A9)} the sum has a fixed number of bounded terms, and
\begin{equation}
\frac{1}{d}\bW^\top\bPsi^{-1}\bW
= \frac{1}{d}\sum_{j \in S}\sigma_j^{-2}\bW_{j\cdot}\bW_{j\cdot}^\top \to \bzero,
\end{equation}
and similarly $\Omega = \bzero$.
Neither limit is positive definite, so \textbf{(A5)} fails.
The simulation design of Section \ref{sec:exp}, in which $10$ features carry signal and the
remaining $d-10$ have zero loadings, is of exactly this form.
Moving forward, we will analyze the parameter estimates of the LFA model: Let $\bx_1, \dots, \bx_n$ be $d-$dimensional $n$ i.i.d. samples from the LFA model $\bx_k = \bW\bgamma_k + \beps_k$. 
For our analysis, we will also consider the following assumptions. 
\begin{itemize}
\item[(\textbf{A7})] $\beps_i \sim \mathcal{N}(\mathbf{0}, \bPsi)$ and $\bPsi = \mathrm{diag}(\sigma_1^2, \sigma_2^2, \ldots, \sigma_d^2)$ 
\item[(\textbf{A8})] $\bgamma_k \sim \mathcal{N}(\mathbf{0}, \mathbf{I}_r)$. The random vectors $\bgamma_k$ and $\beps_{k'}$ are independent for any $\{k\not =k' \in \{1,2,\dots n\}\}$
    
\item[(\textbf{A9})]  The true, underlying parameters of the LFA model are assumed to be well-behaved: $0 < \sigma_{\min}^2 \le \sigma_j^2 \le \sigma_{\max}^2$ for all $j \in \{1, \dots, d\}$.
  
\end{itemize}

Under \textbf{(A8)} the factors are random, and \textbf{(A3)} is satisfied conditionally on
$\{\bgamma_k\}$: by the strong law of large numbers $M_{\Gamma\Gamma} \to \bI_r$ almost surely,
and $M_{\Gamma\Gamma}$ is bounded with probability tending to one, so \textbf{(A9)} holds as well.
The rates of \cite{bai2012statistical} are therefore available conditionally.
The constants they involve depend on $\{\bgamma_k\}$ only through $M_{\Gamma\Gamma}$ and the bound
$C$, both of which are stochastically bounded; the family of conditional laws is consequently
uniformly tight and the conditional rates transfer to unconditional $O_p$ statements , which is what Theorem \ref{thm:scalar_bernstein} requires.
Conditioning also replaces $\bW\bW^\top$ by $\bW M_{\Gamma\Gamma}\bW^\top$ in the covariance; since
$\|M_{\Gamma\Gamma} - \bI_r\| = O_p(1/\sqrt{n})$, the resulting discrepancy is $O_p(1/\sqrt{n})$
and does not affect the rate in Theorem \ref{thm:asym}.

The maximum likelihood estimates of the LFA model satisfy a set of stationarity conditions that
have been known since the classical treatment of factor analysis \cite{lawley1962factor}.
Among them is an exact relation between the estimated signal variance, the estimated noise
variance and the sample variance of each feature.
We restate this relation in our notation in Lemma \ref{lem:lfa-fixedpoint}, and then use it to
decompose the signal-variance estimation error for our purpose.

\begin{lemma}[\textbf{Stationary Point Characterization of the LFA Log-Likelihood}]
\label{lem:lfa-fixedpoint}
Let $(\hat{\bW},\hat{\bPsi})$ be a stationary point of the LFA log-likelihood, and suppose
Assumption \textbf{(A6)} holds, so that $\hat{\bPsi}$ is invertible.
Write $\bM = \hat{\bW}\hat{\bW}^\top + \hat{\bPsi}$, and let $D(\bX)$ denote the diagonal matrix
formed by the diagonal elements of $\bX$.
Then the following two identities hold:
\begin{align}
(\bI - \bM^{-1}\widehat{\bSigma})\bM^{-1}\hat{\bW} &= \bzero, \label{eq:lfa-fp1}\\
D(\bM^{-1}\widehat{\bSigma}) &= \bI. \label{eq:lfa-fp2}
\end{align}
Consequently, for each coordinate $i \in \{1,\dots,d\}$,
\begin{equation}
(\hat{\bW}\hat{\bW}^\top)_{ii} = \widehat{\bSigma}_{ii} - \hat{\bPsi}_{ii}.
\label{eq:lfa-identity}
\end{equation}
\end{lemma}

\noindent
Identities \eqref{eq:lfa-fp1} and \eqref{eq:lfa-fp2} are the first-order conditions obtained by
differentiating the log-likelihood with respect to $\bW$ and with respect to the diagonal of
$\bPsi$, respectively.Identity \eqref{eq:lfa-identity} is classical: it is Eq.\ (11) of \cite{lawley1962factor}, which states that at the maximum likelihood solution the fitted variances reproduce the sample variances exactly, so that the estimated noise variance of a feature is its sample variance minus its estimated communality.
We give a short self-contained derivation in our notation in the Supplementary Material.
Since the EM algorithm of Theorem \ref{thm:lfa} converges to a stationary point of the likelihood
\cite{wu1983convergence}, \eqref{eq:lfa-identity} applies to the associated LFA model parameter estimates.

Our use of \eqref{eq:lfa-identity} is as an error decomposition rather than as a statement about
the fitted model.
Subtracting \eqref{eq:lfa-identity} from the population relation
$(\bW\bW^\top)_{ii} = \bSigma_{ii} - \bPsi_{ii}$ and applying the triangle inequality gives, for
each $i \in \{1,\dots,d\}$,
\begin{equation}
\big|(\bW\bW^\top)_{ii} - (\hat{\bW}\hat{\bW}^\top)_{ii}\big|
\le \big|\widehat{\bSigma}_{ii} - \bSigma_{ii}\big| + \big|\hat{\bPsi}_{ii} - \bPsi_{ii}\big|.
\label{eq:lfa-bound1}
\end{equation}
This is the form in which the identity is useful here.
Estimating $\bW$ and $\bPsi$ jointly is difficult because the EM updates couple them along the
optimization path, and the signal variance $(\bW\bW^\top)_{ii}$ is not a quantity to which
concentration inequalities apply directly.
Inequality \eqref{eq:lfa-bound1} replaces it by two terms that are individually tractable: the
first is the deviation of a single sample variance, controlled by Theorem
\ref{thm:scalar_bernstein}, and the second is the noise variance estimation error.
The decomposition is exact at any stationary point and requires no distributional assumption on
the data, so it holds for finite $n$ and is not contingent on the EM algorithm reaching a global
optimum.
It is the basis for both the asymptotic rate of Theorem \ref{thm:asym} and, under the isotropic
noise assumption, the finite-sample analysis of PPCA.

The problem is now reduced to bounding the sample variance error (which we can do with Theorem \ref{thm:scalar_bernstein}) and the noise variance estimation error. 

We will derive some asymptotic bounds for signals and SNR estimation error in the following theorem.
\begin{theorem}\label{thm:asym}[\textbf{Asymptotic Bounds for Signal Variance and SNR Estimation Error}]
    Under the assumptions (\textbf{A5-A8 and A9(ii)}), when $(n,d) \to \infty$, then 
    \begin{align}
        \abs{ (\bW\bW^T)_{ii} - (\hat{\bW}\hat{\bW}^T)_{ii} } &= O_p(\sqrt{d/n}), \\
        \abs{\widehat{\bSNR}_i - \bSNR_i} &= O_p(\sqrt{d/n}).
    \end{align}
\end{theorem}

\noindent\textbf{An open problem.}
A finite-sample counterpart of Theorem \ref{thm:asym} would require bounding
$|\hat{\bPsi}_{ii} - \bPsi_{ii}|$ at fixed $n$, which we have not been able to obtain.
Two obstacles arise.
First, the EM updates couple $\hat{\bW}$ and $\hat{\bPsi}$ along the optimization path
\cite{rubin1982algorithms}, so neither can be analyzed in isolation.
Second, the natural plug-in estimator of $\sigma_i^2$ evaluates residuals using parameters
estimated from the same data, which introduces a bias that does not vanish at any fixed $n$.
Sample splitting, which restores the independence between the estimated parameters and the data on
which the residuals are evaluated, appears to be a promising route, and we leave this to future
work.


\begin{figure*}[t]
\begin{tabular}{ccc}
\hspace{-3mm}\includegraphics[width = 0.33\linewidth]{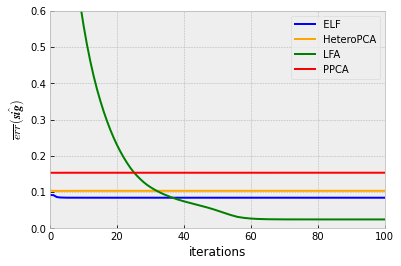}
&\hspace{-3mm}\includegraphics[width = 0.33\linewidth]{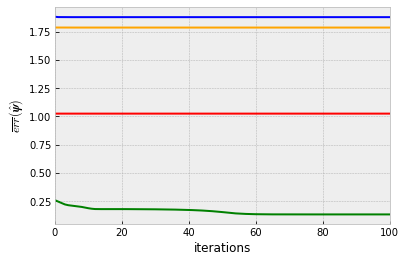}
&\hspace{-3mm}\includegraphics[width=0.33\linewidth]{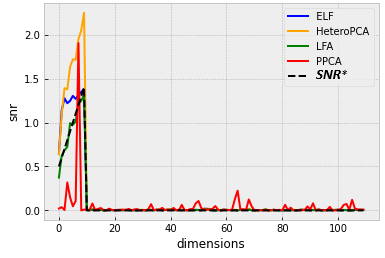}\vspace{-2mm}\\
(a) $\hat{\bsig}$
&(b) $\hat{\bpsi}$
&(c) $\hat{\bSNR}$\vspace{-4mm}
\end{tabular}
\caption{Estimation error vs. several iterations for (a) the signal variance $\hat{\bsig}$, and (b) the noise variance, $\hat{\bpsi}$, when $n=1000$ and $d=110$. 
(c) The true SNR ($\bSNR^{\ast}$) and the estimated SNRs obtained by the four methods.}
\label{fig:itr_dev}
\vspace{-6mm}
\end{figure*}

\section{Multi-class Classification}

The proposed feature selection method can be used for multi-class classification.
For that, each class is represented as a PPCA or LFA model, and the parameters are estimated using one of the four methods described in Section \ref{sec:lrmodels} based solely on the data from that class.
Then, feature selection is performed separately for each class using SNR, as described in Algorithm \ref{algo:snr}.

After selecting the relevant features, the next step involves using these models for multi-class classification by computing the Mahalanobis distance:
\vspace{-3mm}
\begin{equation}\label{eq:MD}
MD(\bx,\bmu,\bSigma)  = (\bx- \bmu)^T \bSigma^{-1}(\bx-\bmu),
\vspace{-3mm}
\end{equation}
of an observation to each class and finding the nearest class.
In high-dimensional scenarios, the Mahalanobis distance is preferred over the Euclidean distance because it considers the covariance structure of the data, enhancing the classification accuracy. 

More exactly, assuming that there are $C$ classes, an observation $\bx$ is classified in two steps:
\begin{enumerate}
\item The Mahalanobis distance $MD_j =MD(\bx^{(j)},\hat{\bmu}_j,\hat{\bSigma}_j)$ is calculated for every class $j \in \{1,2,\cdots,C\}$, where the vector $\bx^{(j)}$ contains the values of the selected features for class $j$, and $\hat{\bmu}_j$ and $\hat{\bSigma}_j$ are the estimated mean and covariance matrix on the selected features from class $j$.
\item The predicted class $k$ is obtained as 
\begin{equation}
k = \argmin_{j \in \{1,2,\cdots,C\}}MD_j.
\end{equation}
\end{enumerate}
When $m$ is large, computing $MD_j$ becomes computationally expensive due to the multiplication with a size $m \times m$ matrix. 
In such cases, we employ an alternative r-score, first introduced by \cite{wang2023scalable} for PPCA, as described in Prop. \ref{Def:rsc}, and generalized for LFA in Theorem \ref{thm:grsc} below.
\begin{proposition}
[due to \cite{wang2023scalable}]\label{Def:rsc}
If $\bSigma$ admits a rank-r eigendecomposition of the form:
\begin{equation}
\label{pca}
          \bSigma = \bL\mathbf{D} \bL^T + \lambda \bI_m,  
\end{equation}


with $\bL \in \RR^{m \times r}$, diagonal $\bD= \diag(\bd) \in \RR^{r \times r}$ with positive entries, and $\lambda>0$, the Mahalanobis distance can be computed as:
\begin{equation}\label{eq:mdscore}
    MD(\bx,\bmu,\bSigma)  = r(\bx;\bmu,\bL,\mathbf{D},\lambda),
\vspace{-3mm}
\end{equation}
where 
\begin{equation}\label{eq:rscore}
r(\bx;\bmu,\bL,\mathbf{D},\lambda) = \|\bx-\bmu\|_2^2 / \lambda - \|\bu(\bx)\|_2^2/ \lambda,
\end{equation}

with $\bu(\bx) = \diag(\frac{\sqrt{\bd}}{\sqrt{\bd+\lambda \bone_r}})\bL^T (\bx - \bmu)$, and $\bone_r = (1,1,\cdots, 1)^T$. 
The operation $\frac{\sqrt{\bd}}{\sqrt{\bd+\lambda \bone_r}}$ is performed element-wise.
\end{proposition}

In this paper, we generalize Prop. \ref{Def:rsc} to the LFA model, as outlined in Theorem \ref{thm:grsc}, also proved in the Supplementary Material. 
Here, $\bPsi$ is a diagonal matrix containing the estimated noise variances on its diagonal, corresponding to the selected features.
\begin{theorem}\label{thm:grsc}
If
\vspace{-3mm}
\begin{equation}
\label{eq:pca1}
    \bSigma = \bL\mathbf{D} \bL^T + \bPsi,    
\end{equation}
with $\bL \in \RR^{m \times r}$, and diagonal matrices $\bD \in \RR^{r \times r}$ and $\bPsi\in \RR^{m \times m}$ with positive entries, the Mahalanobis distance can be computed as:

\begin{equation}\label{eq:rscore1}
    MD(\bx,\bmu,\bSigma)  = r(\bPsi^{-\frac{1}{2}}\bx;\bPsi^{-\frac{1}{2}}\bmu,\bL',\mathbf{D}',1) 
\end{equation}
where $r(\bx;\bmu,\bL,\mathbf{D},\lambda) $ is defined in Eq. \eqref{eq:rscore}, and $\bL^{\prime}_{d \times r}$ and $\bD^{\prime}_{r \times r}$ are obtained by SVD on $\bSigma^{\prime} = \bPsi^{-\frac{1}{2}}\bSigma\bPsi^{-\frac{1}{2}}$.
\end{theorem}

\section{Experiments}\label{sec:exp}
Experiments include one-class simulations that test the parameter estimation and feature selection capabilities of the different methods discussed in Section \ref{sec:lrmodels}, feature selection evaluation on real multi-class datasets, and comparisons with some existing feature selection and class-incremental learning methods.

\subsection{Simulations}

\begin{figure*}[t]
\scalebox{1.05}{
\begin{tabular}{ccc}
\hspace{-3mm}\includegraphics[width = 0.33\linewidth]{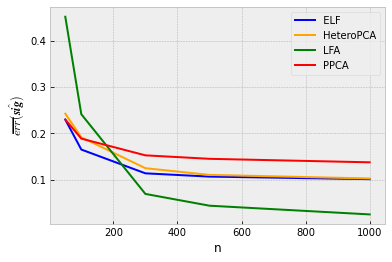}
&\hspace{-3mm}\includegraphics[width = 0.33\linewidth]{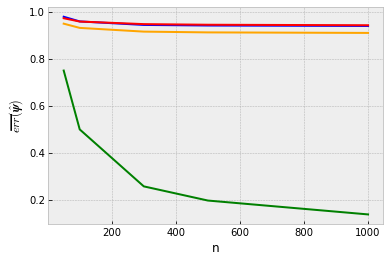}
&\hspace{-3mm}\includegraphics[width=0.33\linewidth]{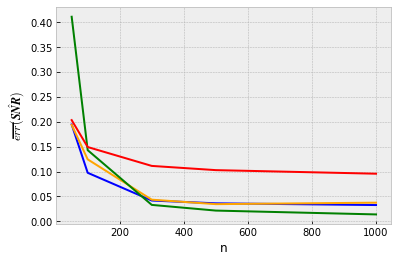}\vspace{-2mm}\\
(a) $\hat{\bsig}$
&(b) $\hat{\bpsi}$
&(c) $\hat{\bSNR}$
\end{tabular}}
\caption{Average estimation error vs. number $n$ of observations for (a) the signal variance, $\hat{\bsig}$, (b) the noise variance, $\hat{\bpsi}$, and (c) the SNRs, $\hat{\bSNR}$, when $d=110$.}
\label{fig:conv_n}
\end{figure*}
The simulated data, $\bX_{n \times d}$, is a concatenation of 10 relevant features, and $d_{noise}=d-10$ irrelevant (noisy) features. 

In our simulation, the true SNR vector is:
\[
\bSNR_i = 
    \begin{cases}
          \frac{1}{2} + \frac{i-1}{10} & i\in \{1,...,10\}\\
          0 & i > 10\\
    \end{cases} 
\]
We generate $\bX = (\bx_1,\bx_2,\cdots,\bx_n)^T,\bx_i\in \RR^d$ using the model \eqref{eq:LFAm}:
\begin{equation}\label{eq:gendata}
    \bx_i = \bW\bgamma_i + \beps_i 
          = \left[ \begin{array}{l}
                    \bW^{(1)}_{10\times r} \vspace{0.2cm}\\
                    \bO_{d_{noise}\times r}
                    \end{array}
             \right]\bgamma_i + \beps_i.
\end{equation}
Here, $\bW^{(1)}_{ij} \overset{\text{i.i.d}}{\sim} \N(0,1)$, and all the elements except in the first $10$ rows are $0$.
The latent vectors $\bgamma_i \overset{\text{i.i.d}}{\sim} \N(\bzero,\bI_r)$ have $r = 3$. 
The noise variable, $\beps_i \overset{\text{i.i.d}}{\sim} \N(\bzero, \bPsi)$ with $\bPsi = \diag(\sigma_1, \sigma_2, \cdots, \sigma_{d}^{ 2})$ and
\[
\sigma_i^{ 2} = 
    \begin{cases}
          \frac{\sum_{j = 1}^{r}\bW_{ij}^2}{\bSNR_i} & i\in \{1,...,10\},\\
           \overset{\text{i.i.d}}{\sim} Uniform(r/1.4,r/0.5) & i > 10.
    \end{cases} 
\]
Smaller SNRs usually correspond to larger error variances. 
The true SNRs range from 0.5 (small) to 1.4 (large) to make true feature recovery more challenging. 
The noise variable variances for the irrelevant dimensions are made comparable to those of the signal dimensions using the uniform distribution, as specified above. 

\noindent{\bf Parameter Estimation Evaluation.}
We evaluate the estimation process by comparing the estimation error associated with $(\hat{\bsig},\hat{\bpsi}, \hat{\bSNR})$ for the employed low-rank generative methods.  
The corresponding true values of the parameters are $(\bsig,\bpsi,\bSNR)$.  

An example of the estimation errors vs. the number of iterations for one dataset is shown in Figure \ref{fig:itr_dev}(a) and (b), and the obtained SNRs, along with the true SNRs, are shown in Figure \ref{fig:itr_dev} (c). 
The estimation errors of the LFA method are the best, given that there are sufficiently many iterations. 
The performance of ELF and HeteroPCA methods are similar. 
ELF exhibits slightly superior performance over HeteroPCA in estimating the signal ($\bsig$), while HeteroPCA outperforms ELF in estimating $\bpsi$. 
The PPCA method has the largest estimation error for $\hat{\bsig}$, and the error for $\hat{\bpsi}$ is smaller than the ELF and HeteroPCA methods.

Figure \ref{fig:itr_dev}(c) displays the estimated SNRs, $\hat{\bSNR}$, alongside the true SNRs, $\bSNR$. All methods, except PPCA, estimate close to $0$ values for the noisy features $(11-110)$. PPCA often overestimates $\bSNR$s for noise, with some signals having near-zero $\hat{\bSNR}$s. LFA provides the most accurate estimates. ELF's $\hat{\bSNR}$ values are close to the $\bSNR$ but marginally less aligned than LFA’s. However, HeteroPCA overestimates positive $\bSNR$ values the most but perfectly captures the pattern.

For a more thorough evaluation, we compute the average of mean absolute deviation (MAD) over $R=50$ independent runs, defined as 
\begin{equation}
\overline{err}(\hat{\btheta}) = \frac{1}{R}\sum_{r=1}^R(\frac{1}{d}\sum_{j=1}^d |\btheta^{ r}_j - \hat{\btheta}^r_j|),
\end{equation}
where $\hat{\btheta}_j^r$ is the estimate of the true parameter $\btheta^{ r}_j$, where $\hat{\btheta}$ could be one of $(\bsig,\bpsi,\bSNR)$. 


The average MAD values for data with $d=110$ and different sample sizes are shown in Figure \ref{fig:conv_n}. 
From Figure \ref{fig:conv_n}, it is clear that the LFA method obtains the smallest estimation errors across all parameters for large $n$. 
ELF and HeteroPCA show similar performance, with ELF slightly better for signal estimation and HeteroPCA better for noise. 
LFA, HeteroPCA, and ELF show comparable performance for SNR estimation. 
The performance of PPCA lags behind in the estimation process as it assumes homoscedastic noise.
\begin{figure}
    \includegraphics[height = 0.5\linewidth]{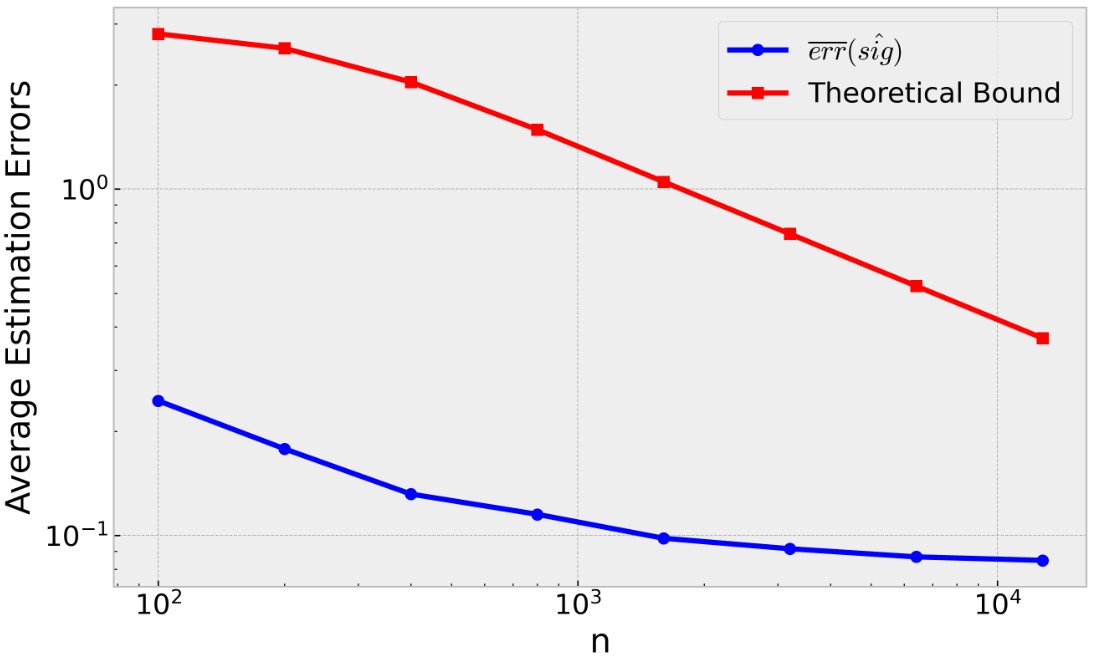}
    \caption{Estimation error and theoretical bound vs. number of observations ($n$) for the PPCA signal variance.}
    \label{fig:sim-ppca}
\end{figure}

Figure \ref{fig:sim-ppca} provides a comprehensive empirical validation of our theoretical analysis. 
Figure \ref{fig:sim-ppca} showcases the comparison between  $\overline{err}(\hat{\bsig})$ and the corresponding theoretical bound vs $n$ on a log-log scale for PPCA. 
This figure empirically validates our non-asymptotic analysis for the signal variance estimator, $\bsig$. 

The near-linear slope of both curves on the log-log scale visually demonstrates the expected $1/\sqrt{n}$ rate of convergence.


\begin{table}[t]
\centering
\caption{Feature selection accuracy (\%) for simulated data.}
    \label{tab:Sim1}
\scalebox{0.99}{
\begin{tabular}{|c|c| >{\centering\arraybackslash}m{1cm}| >{\centering\arraybackslash}m{1cm}| >{\centering\arraybackslash}m{1cm}| >{\centering\arraybackslash}m{1cm}|}
    \hline
    \textbf{Method} & \diagbox[]{Noise}{\vspace{-0.2cm} n\hspace{0.4cm}} & 50 & 100 & 300 & 1000 \\
    \hline
    \multirow{3}{*}{PPCA} & 10 & 71.2 & 82.4 & 88 & 95.6 \\
                                   & 50 & 55.4 & 72.6 & 79.8 & 91.4 \\
                                   & 100 & 49 & 62.8 & 82 & 90 \\
    \hline
    \multirow{3}{*}{LFA}  & 10 & \textbf{90.6} & \textbf{97} & \textbf{100} & \textbf{100} \\
                                   & 50 & 70.4 & 91.8 & \textbf{100} & \textbf{100} \\
                                   & 100 & 57.4 & \textbf{87.4} & \textbf{99.4} & \textbf{100} \\
    \hline
    \multirow{3}{*}{ELF}  & 10 & 87.6 & 94 & 98 & \textbf{100}\\
                                   & 50 & \textbf{73.4} & \textbf{92.8} & 98.6 & 99.8 \\
                                   & 100 & \textbf{58} & 87.2 & \textbf{99.4} & 99.6 \\
    \hline
    \multirow{3}{*}{HeteroPCA} & 10 & 84.4 & 93 & 98.8 & 99.5 \\
                                        & 50 & 65.6 & 87 & 94.4 & 99.2 \\
                                        & 100 & 55.8 & 75.6 & 96.4 & 99\\
    \hline
\end{tabular}
}
\end{table}

\noindent{\bf Feature Recovery Evaluation.}
In this experiment, we consider feature selection accuracy: a pivotal aspect of method assessment. 
To measure how accurate the feature selection process is, we compare the set of indices of the features that are truly relevant (signals), denoted as $\mathscr{I}_{true}$ with the ones each method has predicted as relevant, denoted as $\mathscr{I}_{pred}$.
The feature selection accuracy is defined as the average percentage of features that are correctly recovered:

\begin{equation}
Acc = E(|\mathscr{I}_{true} \cap \mathscr{I}_{pred}|)/|\mathscr{I}_{true}|.
\end{equation}
In our experiments, the expected value is computed over 50 independent runs and the number of selected features was $|\mathscr{I}_{pred}|=|\mathscr{I}_{true}|$.
The $Acc$ values  for different combinations of $n$ and $d_{noise}$ are shown in Table \ref{tab:Sim1}.
As $d_{noise}$ increases, identifying relevant features becomes more challenging across all methods. 
Adequate data, indicated by a larger $n$, becomes necessary in such scenarios. 
LFA consistently identifies the true features, especially for moderate $n$ values like $n=300$, with ELF performing comparably. Both methods surpass the performance of HeteroPCA and PPCA.

\subsection{Real Data Experiments}
To complement the preceding theoretical analysis and demonstrate the practical efficacy of our proposed Signal-to-Noise Ratio (SNR) based feature selection methods, we conduct a comprehensive empirical study. 
While our theoretical results provide a rigorous foundation for the reliability of the LFA model, this section aims to validate their performance in real-world, high-dimensional classification tasks. 
We evaluate our family of latent factor models---PPCA, LFA, HeteroPCA, and ELF---on three widely-utilized and challenging image classification benchmarks: CIFAR-10\cite{cifar10}, CIFAR-100\cite{cifar100}, and the large-scale ImageNet-1k dataset\cite{ILSVRC15}. 
These datasets provide a robust testbed, spanning a range of complexities from 10 to 1,000 classes.

Recognizing that the quality of the initial feature representation is paramount, we leverage a state-of-the-art foundation model, \textbf{DINOv3} \cite{simeoni2025DINOv3}, to extract high-dimensional embeddings from the raw images. 
It is a self-supervised visual representation learning model trained on the massive LVD-1.7B dataset, a curated collection of nearly 1.7 billion images. 
DINOv3 learns powerful representations from images alone via a sophisticated image token-matching and self-distillation process. 
The model's architecture is a large-scale Vision Transformer (ViT-L) that processes images by dividing them into a grid of non-overlapping $16 \times 16$ pixel patches.

For our purposes, we used the dinov3-vitl16-pretrain-lvd1689m classifier. 
The DINOv3 feature extractor is employed with medium resolution $512 \times 512$ images. 
Features of dimension $d=1024$ were extracted after the classifier's attention pooling layer for all the datasets.

The performance of our methods is benchmarked against two established and competitive techniques: Feature Selection with Annealing (FSA) \cite{barbu2016feature} and the Thresholding-based Iterative Selection Procedure (TISP) with the hard thresholding penalty \cite{she2012iterative}. 
FSA enforces sparsity through an annealing schedule based on coefficient magnitudes, while TISP utilizes iterative hard-thresholding rules. 
Both baselines are implemented via a single-layer neural network optimized with cross-entropy loss and the Adam optimizer \cite{kingma2014adam}.

\noindent \textbf{Experimental Protocol and Evaluation.} Experiments were conducted on an 11th Gen Intel octa-core processor. 
Performance is assessed across dual axes: downstream classification accuracy as a function of feature subset size and total computational overhead, ensuring a robust evaluation of both discriminative power and scalability.
\begin{table}[t]
\centering
\caption{Training time (seconds) for FSA and TISP on the datasets evaluated.}
\label{tab:time_c}
\vskip -3mm
\scalebox{1.}{
\begin{tabular}{|c|c|c|c|c|c|c|c|}
\hline
Dataset & Method &\multicolumn{6}{c|}{\# Selected Features} \\
\cline{3-8}
& &1024 & 1000 & 800 & 600 & 400 & 200\\
\hline
\multirow{2}{*}{CIFAR-10} & FSA  &24 & 23 & 23 & 22 & 22 & 21  \\
         &TISP & 26 & 26 & 26 & 25 & 25 & 24   \\
\hline
\hline
\multirow{2}{*}{CIFAR-100} & FSA  &27 & 26 & 26 & 26 &25 & 25 \\
         &TISP  &30 & 29 & 28 & 27 & 26 & 26 \\
\hline
\hline
& &1024 & 1000 & 800 & 600 & 400 & 200\\
\hline
\multirow{2}{*}{ImageNet} & FSA &3963 &3786 &2479 &1222 &1024 &1018 \\
         &TISP & 3963 &3863 &2326 &1136 & 1034& 1022 \\
\hline
\end{tabular}
}
\vspace{-3mm}
\end{table}

\begin{table}[ht]
\centering
\caption{Training time (seconds) for low-rank generative methods on the datasets evaluated.}\label{tab:time_l}
\vspace{-3mm}
\scalebox{1.}{
\begin{tabular}{|c|c|c|c|c|c|}
\hline
Dataset & \# Features & \multicolumn{4}{c|}{\textbf{Methods}} \\
\cline{3-6}
& & \small{PPCA} & \small{LFA}& \small{ELF}& \small{HeteroPCA}\\
\hline
CIFAR-10 & 1024  &  5 & 18 & 9 &57 \\
\hline
CIFAR-100 & 1024 & 9 & 24 & 12 & 59\\
\hline
ImageNet & 1024 &  133 &476 &916 &611 \\
\hline
\end{tabular}}
\vspace{-4mm}
\end{table}

Training efficiencies are contrasted in Tables \ref{tab:time_c} and \ref{tab:time_l}, highlighting that our class-specific SNR method facilitates multi-level feature selection without the retraining necessitated by embedded approaches. 

\begin{table}[t]
\centering
\caption{Classification accuracy ($\%$) for different methods on real datasets.}
\vspace{-4mm}
\scalebox{1.2}{
\begin{tabular}{|c|c|c|c|c|c|c|}
\hline
Method & \multicolumn{6}{c|}{\# Selected Features} \\
\hline
\multicolumn{6}{l}{
CIFAR-10, $n=60,000,d=1024$}\\
\hline
 &1024 & 1000 & 800 & 600 & 400 & 200\\
\hline
 \small{FSA}  & 98.87& 98.93 & 98.84 & 98.79 & 97.67 & 94.24\\
\small{TISP} &  \textbf{98.93} & \textbf{98.93} & 98.78 & 98.7 & 97.78 & 94.53\\
\small{ELF}  &  98.88 & 98.9 & 98.92 & 98.84 & 98.74 & 98.08\\
\small{HeteroPCA} & 98.88 & \textbf{98.93} & 98.91 & 98.86 & \textbf{98.83} & \textbf{98.36}\\
\small{LFA} &98.88 & \textbf{98.93} & \textbf{98.94} & 98.87 & 98.78 & 98.2\\
\small{PPCA} & 98.88 & 98.91 & 98.91 & \textbf{98.91} & 98.81 & \textbf{98.36}\\
\hline
\multicolumn{6}{l}{
CIFAR-100, $n=60,000,d=1024$}\\
\hline
 &1024 & 1000 & 800 & 600 & 400 & 200\\
\hline
 \small{FSA} &91.74 & 91.31 & 91 & 90.72 & 88.93 & 83.68\\
\small{TISP} &91.68 & 91.61 & 91.47 & 90.48 & 88.93 & 82.68\\
\small{ELF} &\textbf{91.8} & 91.8 & 91.8 & 91.52 & 91.25 & 89.67\\
\small{HeteroPCA} &\textbf{91.8} & \textbf{91.87} & 91.77 & \textbf{91.76} & 91.25 & 89.75\\
\small{LFA}& \textbf{91.8} & 91.8 & \textbf{92}& 91.51 & 91.18 & 89.17\\
\small{PPCA} & \textbf{91.8} & 91.8 & 91.82 & 91.7 & \textbf{91.4} & \textbf{90.36}\\
\hline
\multicolumn{6}{l}{
ImageNet, $n=1.2$ million, $d=1024$}\\
\hline
 &1024 & 1000 & 800 & 600 & 400 & 200 \\
\hline
 \small{FSA} &83.05 & 82.8 & 82 & 78.93 & 70.35 & 46 \\
\small{TISP}  & 83.05 &82.13 & 81.67 & 78.67 & 71.3 & 45.01\\
\small{ELF} & \textbf{83.93} & 83.92 & 83.92 & 83.75 & 83.1 & 80.76 \\
\small{HeteroPCA} & \textbf{83.93} & 83.92 & 83.92 & 83.73 & 83.13 & 80.71\\
LFA  &\textbf{83.93} & 83.93 & 83.87 & 83.66 & 83.13 & 80.78\\
\small{PPCA}& \textbf{83.93} &\textbf{ 83.94} & \textbf{84.03} & \textbf{83.91} & \textbf{83.76} & \textbf{81.7} \\
\hline
\end{tabular}
}
\label{tab:comp}
\end{table}

\begin{figure}[ht]
\includegraphics[width = \linewidth]{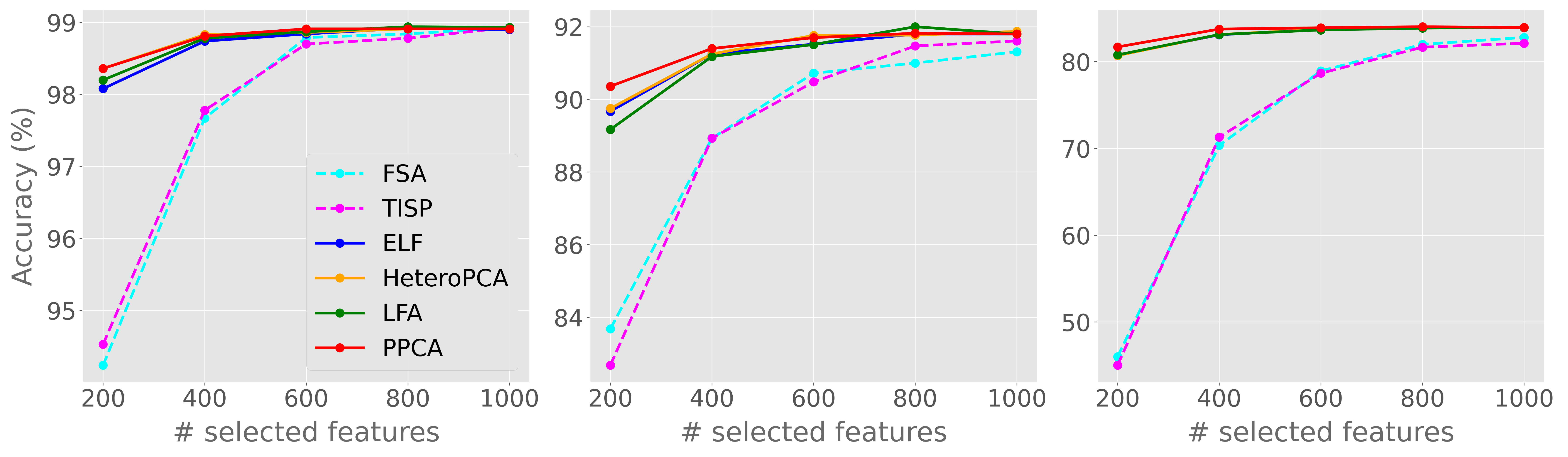}
\caption{Accuracy vs. number of selected features for CIFAR-10 (left), CIFAR-100(middle)  and ImageNet (right).}
\label{fig:acc}
\end{figure}

Figure \ref{fig:acc} and Table \ref{tab:comp} present classification accuracy (in percent) for our proposed generative feature selection methods (ELF, HeteroPCA, LFA, PPCA) and the two discriminative benchmarks (FSA, TISP). 
The evaluation is performed using all 1024 features, then reduced to a highly compressed subset of 200. For PPCA and LFA-based methods, the rank is 20.
A detailed analysis of these results reveals several consistent and significant trends.

\noindent \textbf{Performance on CIFAR-10.} Using DINOv3 features, all methods achieve a high baseline accuracy of approximately $98.9\%$. In the high-dimensional regime ($\geq 800$ features), performance is comparable across all models. However, at a reduced dimensionality of 200 features, a significant divergence occurs: discriminative baselines (FSA and TISP) exhibit a marked decline to approximately $94\%$, while our generative SNR-based methods (LFA, HeteroPCA, and PPCA) consistently maintain accuracies above $98\%$.

\noindent \textbf{Performance on CIFAR-100.} The complexity of the 100-class CIFAR-100 task amplifies these observed trends. While the full-feature baseline is approximately $91.8\%$, aggressive pruning to 200 features causes FSA and TISP accuracies to plummet by over eight percentage points to $\sim 83\%$. Conversely, generative methods exhibit exceptional stability; PPCA retains an accuracy of $90.36\%$, representing a marginal $1.4$-point degradation. Notably, LFA achieves the highest peak accuracy ($92\%$ at 800 features), suggesting that SNR-based selection actively enhances performance by filtering idiosyncratic noise.

\noindent \textbf{Performance on ImageNet-1k.} The large-scale ImageNet-1k dataset highlights a fundamental collapse in discriminative pruning strategies, as FSA and TISP accuracies fall to $\sim 45\%$ when reduced to 200 features. In stark contrast, all four generative methods remain robust, preserving accuracies above $80\%$. PPCA is the standout performer, achieving a peak accuracy of $84.03\%$ and retaining $81.7\%$ accuracy even after an $80\%$ reduction in dimensionality. These results provide strong empirical validation for our central thesis: the SNR criterion derived from generative models identifies the most stable and informative features, ensuring reliability and scalability in high-dimensional, complex classification tasks.

\subsection{Class-Incremental Learning Experiments}
Our proposed framework aligns with the modern school of thought in Class-Incremental Learning, leveraging powerful pre-trained Models (PTMs) to provide a stable feature space. 
However, it introduces a critical and novel extension by integrating a class-specific feature selection mechanism. 
Building on the philosophy of works such as \cite{wang2023scalable}, which use a frozen feature extractor and separate generative models per class, our approach not only learns a unique distribution for each class but also identifies the most salient original features necessary for its recognition.
This architecture offers a fundamental and structural solution to catastrophic forgetting, contrasting sharply with the preservation and bias-correction methods discussed previously. 
Unlike regularization-based approaches \cite{hinton2015distilling, rebuffi2017icarl} that require exemplars or knowledge distillation to approximate past knowledge, our method perfectly preserves it. 
It also sidesteps the need for complex bias-correction layers \cite{hou2019learning, wu2019large} that re-balance a shared classifier.
The suitability of our framework for CIL is rooted in its class-specific design for the following reasons:
\begin{itemize}
\item \textbf{Structural Immunity to Forgetting:} The core challenge of catastrophic forgetting arises from the overwriting of shared parameters in a monolithic model. 
By dedicating a separate, independent generative model to each class, we eliminate this issue by design. 
The parameters learned for a new class have no architectural pathway to interfere with the parameters of previously learned classes. 
Preserving these learned parameters ensures prior knowledge is fully retained, not just approximated.

\item \textbf{Constant-Time Model Expansion:} Traditional classifiers often require retraining on a growing dataset, with complexity scaling as $O(C)$ with the number of classes $C$. 
Our approach is significantly more scalable. 
The model for each class is "wrapped" tightly around its own observations. 
When a new class is introduced, we train a new, independent model for it. 
This allows the system to expand its knowledge with a computational cost that scales as $O(1)$ per new class, making it ideal for dynamic, large-scale environments.

\item \textbf{Adaptive and Interpretable Feature Selection:} The key novelty of our extension is the integration of Signal-to-Noise Ratio (SNR)-based feature selection. 
As each new class model is trained, we also identify the most discriminative subset of features from the PTM's output space for that specific class. 
This ensures that the system learns not only what a new class looks like (its distribution) but also which features matter most for identifying it. 
This adds a layer of adaptability and interpretability that is not present in prior generative CIL work.
\end{itemize}
While a comprehensive methodological exploration of CIL is beyond the primary scope of this paper, the inherent properties of our framework make it a powerful and elegant solution to the CIL problem. 
We apply it in a CIL setting primarily to empirically validate its structural robustness against catastrophic forgetting.

To validate this property, we adopt the standard CIL protocol from~\cite{hou2019learning}, a widely used benchmark for evaluating incremental methods~\cite{wang2023scalable,zhou2025revisiting,nakata2022revisiting}. 
The protocol proceeds as follows: the model is initialized on a random subset comprising half the dataset's classes (e.g., 50 for CIFAR-100, 500 for ImageNet-1K). 
The remaining classes are divided into 5 or 10 equal incremental tasks, introduced sequentially. 
At each step $b$:
\begin{itemize}
\item Independent generative models are trained for the new classes in the task $b$.
\item The SNR-based feature selection is applied to identify the top discriminative features for these classes.
\item The selected set of features is added to the existing feature pool, and the selected set of features for every class is updated.
\end{itemize}

Evaluation occurs after each step on a held-out test set encompassing all classes seen thus far, using only the selected features for classification via the Mahalanobis distance. 
No exemplars from prior classes are stored or replayed, aligning with the exemplar-free setting \cite{zhou2025revisiting}. 
To ensure robustness, we repeated the process over 5 independent runs with different random seeds for class splitting and report the \emph{average incremental accuracy}: the mean accuracy across all evaluation steps.

\begin{table}[ht]
\caption{Comparison of average incremental accuracy (\%) and final accuracy on CIFAR-100 and ImageNet-1K in the class-incremental learning setting (exemplar-free).}
\label{tab:cil-full}
\scalebox{0.9}{
\begin{tabular}{l|cc|cc|cc|cc} 
\toprule
\multirow{2}{*}{\textbf{Method}} & \multicolumn{4}{c|}{\textbf{CIFAR-100}} & \multicolumn{4}{c}{\textbf{ImageNet-1K}} \\ 
\cmidrule(lr){2-5} \cmidrule(lr){6-9}
& $\bar{a}_{10}$ & $a_{10}$ & $\bar{a}_5$ & $a_{5}$ & $\bar{a}_{10}$ & $a_{10}$ & $\bar{a}_5$ & $a_{5}$ \\ 
\midrule
iCaRL(2017)~\cite{rebuffi2017icarl} & 52.57 & 50 & 57.17 & 45.5 & 46.72 & 45.6 & 51.36 & 39.89 \\
BiC(2019)~\cite{wu2019large} & 53.21 & - & 56.86 & 40.21 & \textbf{84.02} & 73.2 & - & -\\
UCIR (NME)(2019)~\cite{hou2019learning} & 60.12 & - & 63.12 & - & 59.92 & - & 61.56 & - \\
UCIR (CNN)(2019)~\cite{hou2019learning} & 60.18 & 43.39 & 63.42 & - & 61.28 & - & 64.34 & - \\
PODNet(2020)~\cite{douillard2020podnet} & 63.19 & 41.05 & 64.83 & - & 64.13 & - & 66.95 & - \\
PPCA-CLIP(2023)\cite{wang2023scalable} & 69.71 & 72.81 & 69.71 & 72.81 & 71.25 &{\bf 73.73} & 71.25 & \textbf{73.73} \\
LwF (2018)~\cite{li2018learning} & 82.88 & 77.57 & 88.10 & 84.28 & - & - & - & - \\
L2P (2022)~\cite{wang2022learning} & 89.48 & 84.47 & 91.02 & 86.27 & - & - & - & - \\
DualPrompt (2022)~\cite{wang2022dualprompt} & 88.86 & 84.23 & 89.78 & 84.76 & - & - & - & - \\
ACIL (2022)~\cite{ye2022acil} & 91.96 & 90.33 & \textbf{94.00} & 90.73 & - & - & - & - \\
CODA-Prompt (2023)~\cite{smith2023coda} & 91.19 & 87.24 & 92.20 & 88.67 & - & - & - & - \\
LAE (2023)~\cite{liu2023lae} & 86.97 & 81.13 & 88.50 & 82.76 & - & - & - & - \\
DS-AL (2024)~\cite{chen2024dsal} & 83.50 & 86.05 & 88.82 & 85.91 & - & - & - & - \\
SimpleCIL (2024)~\cite{zhou2025revisiting} & 82.31 & 76.21 & 81.12 & 76.21 & - & - & - & - \\
Aper (2024)~\cite{zhou2025revisiting} & 90.91 & 85.81 & 91.56 & 87.51 & - & - & - & - \\
EASE (2024)~\cite{li2024ease} & 92.01 & 87.25 & 92.81 & 89.22 & - & - & - & - \\
\midrule
CLIP(ELF+SNR),640/640 features & - & - & - & - & 76.0 & \textbf{73.73} & \textbf{76.0} & \textbf{73.73} \\
CLIP(ELF+SNR),1750/2560 & 77.37 & 72.81 & 77.37 & 72.81 & - & - & - & - \\
Dinov3(LFA+SNR),800/1024 & \textbf{93.44} & \textbf{92.00} & 93.44 & \textbf{92.00} & - & - & - & - \\
\bottomrule
\end{tabular}
}
\vspace{-4mm}
\end{table}
Table \ref{tab:cil-full} provides a comprehensive and detailed comparison of our proposed frameworks against a wide array of state-of-the-art (SOTA) methods in the challenging \textbf{exemplar-free class-incremental learning setting}. 

The table reports two distinct performance metrics for the CIFAR-100 and the large-scale ImageNet-1K benchmarks: the \textbf{average incremental accuracy} ($\bar{a}$), which is the average performance across all incremental steps, and the \textbf{final accuracy} ($a$), which is the performance at the final step on all classes seen so far. 
These metrics are evaluated for both 5-step and 10-step incremental learning scenarios to assess performance under different learning granularities. 
We employed the Contrastive Language-Image Pre-training (CLIP) \cite{radford2021clip} as a feature extractor alongside Dinov3 to extract features from both datasets and performed class-incremental learning separately using the resulting feature sets for each dataset. 
The CLIP includes two parallel encoders: an image encoder (typically a Vision Transformer or a ResNet) and a text encoder (a standard text Transformer). 
For our purposes, we used a pretrained, modified ResNet-50 classifier, RN50x4, from the CLIP GitHub package \cite{radford2021clip}. 
The CLIP feature extractor is trained on medium-resolution $288 \times 288$ images. 
Accordingly, the input images were resized to $288 \times 288$ for the ImaeNet-1k dataset. 
For CIFAR-100, we resize the original images to $144\times 144$, as resizing to $288 \times 288$ would produce very blurred images. 
Therefore, the feature vectors are 640-dimensional for ImageNet-1K, and 2560-dimensional for the CIFAR datasets, when CLIP is used.

An analysis of historical and recent SOTA methods reveals a clear upward trend in performance over time, particularly on the \textbf{CIFAR-100} benchmark. 
Early methods like iCaRL and BiC established baselines in the 50-60\% accuracy range. 
The field saw significant improvement with the advent of prompt-based learning methods designed for large PTMs, such as L2P, DualPrompt, and CODA-Prompt, which consistently pushed performance to around 90\%. 
As of the latest results, the top-performing SOTA methods in this comparison are ACIL and EASE, with ACIL achieving the highest average accuracy ($\bar{a}_5=94.00\%$) and EASE achieving a strong final accuracy ($a_5=89.22\%$).

Against this highly competitive landscape, our `Dinov3-(LFA+SNR)` framework demonstrates a significant performance leap, establishing a new state-of-the-art on CIFAR-100. 
Our method achieves a final accuracy ($a_5$) of \textbf{92.00\%}, substantially outperforming the previous top method, EASE, by nearly 3 percentage points. 
Furthermore, our average accuracy ($\bar{a}_5$) of \textbf{93.44\%} is highly competitive with the best reported result. 
Critically, this superior performance is achieved not with the full feature set, but with a \textbf{reduced subset of 800 out of 1024 features}, as determined by our SNR criterion. 
This simultaneously validates our feature selection methodology and demonstrates its ability to improve accuracy by removing noisy or redundant information. 

On the more challenging, large-scale \textbf{ImageNet-1K} benchmark, the field is less crowded, as many recent methods do not report results for this task due to overlap between PTM training sets and ImageNet-1K. 
The strongest baseline in this comparison is PPCA-CLIP, which achieves an average accuracy of 71.25\% and a final accuracy of 73.73\%. 
Our `CLIP-(ELF+SNR)` method clearly surpasses this baseline, achieving an average accuracy ($\bar{a}_5$) of \textbf{76\%}.
This represents a significant improvement of nearly 5 percentage points in overall learning stability throughout the incremental process. 
The final accuracy ($a_5$) of 73.73\% matches the baseline, confirming that our method maintains top-tier final performance while demonstrating substantially better learning dynamics. 

In summary, the results unequivocally establish the superiority of our proposed frameworks. 
By synergistically combining powerful pre-trained features, robust generative modeling (LFA/ELF), and principled SNR-based feature selection, our methods advance the state of the art in exemplar-free incremental learning across both standard and large-scale benchmarks.

\section{Conclusion}
\label{sec:conclusion}
This paper presented a feature selection method that treats each class separately.
Every class is represented by a low-rank generative model, and its features are ranked by a signal-to-noise ratio derived from the estimated parameters, which compares the variance explained by the latent factors to the unexplained noise variance.
The criterion is computed once per class, and the number of selected features can be changed by re-thresholding, without refitting the model.
We instantiated the criterion with four estimators, PPCA, LFA, HeteroPCA, and ELF, the last of which is a nonparametric latent factor estimator introduced here, and we generalized the $r$-score of \cite{wang2023scalable} from isotropic to diagonal noise so that Mahalanobis scoring remains efficient for LFA-type models.

Experiments on CIFAR-10, CIFAR-100, and ImageNet-1k with DINOv3 features showed that classification accuracy is largely preserved as the per-class feature subsets are reduced, whereas the accuracy of the two linear-model selectors used for comparison degrades considerably at small subset sizes.
In the class-incremental setting, the method is exemplar-free and its accuracy does not depend on the number of increments into which the classes are divided, which is a direct consequence of the class-specific design: the model and the selected features of a class are never modified after that class has been learned.
The accuracies obtained are competitive with recent methods on both benchmarks.
We note that these comparisons use a stronger pre-trained backbone than several of the methods listed, so the differences reported should be read as evidence that the framework makes effective use of such representations, rather than as an isolated measure of the selection criterion itself.

Two limitations point to the natural continuations of this work.
First, our finite-sample analysis covers the signal variance under PPCA.
The corresponding bound for LFA is harder to obtain, because the EM estimates of the loadings and the noise variances are coupled along the optimization path, and because the plug-in estimator of the noise variance carries a bias that does not vanish at any fixed sample size.
Sample splitting appears to be a promising route, since it restores the independence between the estimated parameters and the data on which the residuals are evaluated, and we intend to pursue it.
Extending the analysis to a non-asymptotic guarantee of correct feature recovery, in the spirit of the asymptotic result of \cite{kansabanik2025feature}, would then follow.
Second, the criterion evaluates a feature by how well the generative model of a single class accounts for it, and therefore does not use information about the other classes.
A feature whose latent structure is shared across many classes will receive a high score for each of them while contributing little to their separation.
This did not prove to be a practical obstacle with the pre-trained representations used here, but a criterion that retains the modularity of the present approach while incorporating a measure of between-class separation is worth investigating.

{
    \bibliographystyle{IEEEtran}
    \bibliography{main}

@String(PAMI = {IEEE Trans. Pattern Anal. Mach. Intell.})

@String(IJCV = {Int. J. Comput. Vis.})

@String(CVPR= {IEEE Conf. Comput. Vis. Pattern Recog.})

@String(ICCV= {Int. Conf. Comput. Vis.})

@String(ECCV= {Eur. Conf. Comput. Vis.})

@String(AAAI = {AAAI})

@String(PAMI  = {IEEE TPAMI})

@String(IJCV  = {IJCV})

@String(CVPR  = {CVPR})

@String(ICCV  = {ICCV})

@String(ECCV  = {ECCV})

@INPROCEEDINGS{7727221,
  author={Antoniades, Andreas and Clive Cheong Took},
  booktitle={IJCNN}, 
  title={Speeding up feature selection: A deep-inspired network pruning algorithm}, 
  year={2016},
  volume={},
  number={},
  pages={360-366},
  doi={10.1109/IJCNN.2016.7727221}}

@INPROCEEDINGS{7486536,
  author={Abbas, Syed Tabish and Sivaswamy, Jayanthi},
  booktitle={ACPR}, 
  title={Latent factor model based classification for detecting abnormalities in retinal images}, 
  year={2015},
  volume={},
  number={},
  pages={411-415},
  doi={10.1109/ACPR.2015.7486536}}

@article{adamczak2011sharp,
  title={Sharp bounds on the rate of convergence of the empirical covariance matrix},
  author={Adamczak, Rados{\l}aw and Litvak, Alexander E and Pajor, Alain and Tomczak-Jaegermann, Nicole},
  journal={Comptes Rendus. Math{\'e}matique},
  volume={349},
  number={3-4},
  pages={195--200},
  year={2011}
}

@article{ahmad2015feature,
  title={Feature selection using particle swarm optimization in intrusion detection},
  author={Ahmad, Iftikhar},
  journal={International Journal of Distributed Sensor Networks},
  volume={11},
  number={10},
  pages={806954},
  year={2015},
  publisher={SAGE Publications Sage UK: London, England}
}

@article{alomari2022hybrid,
  title={Hybrid feature selection based on principal component analysis and grey wolf optimizer algorithm for Arabic news article classification},
  author={Alomari, Osama Ahmad and Elnagar, Ashraf and Afyouni, Imad and Shahin, Ismail and Nassif, Ali Bou and Hashem, Ibrahim Abaker and Tubishat, Mohammad},
  journal={IEEE Access},
  volume={10},
  pages={121816--121830},
  year={2022},
  publisher={IEEE}
}

@article{anderson1963asymptotic,
  title={Asymptotic theory for principal component analysis},
  author={Anderson, Theodore Wilbur},
  journal={Ann. Math. Stat.},
  volume={34},
  number={1},
  pages={122--148},
  year={1963},
  publisher={Institute of Mathematical Statistics}
}

@phdthesis{aziz2023sparse,
  title={Sparse Methods for Latent Class Analysis, Principal Component Analysis and Regression with Missing Data},
  author={Aziz, Rashad},
  year={2023},
  school={The Florida State University}
}

@article{baik2006eigenvalues,
  title={Eigenvalues of large sample covariance matrices of spiked population models},
  author={Baik, Jinho and Silverstein, Jack W},
  journal={Journal of multivariate analysis},
  volume={97},
  number={6},
  pages={1382--1408},
  year={2006},
  publisher={Elsevier}
}

@article{bai2012statistical,
  title={STATISTICAL ANALYSIS OF FACTOR MODELS OF HIGH DIMENSION},
  author={Bai, Jushan and Li, Kunpeng},
  journal={Ann. Stat.},
  pages={436--465},
  year={2012},
  publisher={JSTOR}
}

@article{barbu2016feature,
  title={Feature selection with annealing for computer vision and big data learning},
  author={Barbu, Adrian and She, Yiyuan and Ding, Liangjing and Gramajo, Gary},
  journal={IEEE Trans. on PAMI},
  volume={39},
  number={2},
  pages={272--286},
  year={2017}
}

@inproceedings{bai2008central,
  title={Central limit theorems for eigenvalues in a spiked population model},
  author={Bai, Zhidong and Yao, Jian-feng},
  booktitle={Annales de l'IHP Probabilit{\'e}s et statistiques},
  volume={44},
  pages={447--474},
  year={2008}
}

@article{bihl2025physics,
  title={Physics-Informed Feature Engineering and R2-Based Signal-to-Noise Ratio Feature Selection to Predict Concrete Shear Strength},
  author={Bihl, Trevor J and Young II, William A and Moyer, Adam},
  journal={Mathematics},
  year={2025},
  publisher={Multidisciplinary Digital Publishing Institute}
}

@inproceedings{boutsidis2008unsupervised,
  title={Unsupervised feature selection for principal components analysis},
  author={Boutsidis, Christos and Mahoney, Michael W and Drineas, Petros},
  booktitle={Proceedings of the 14th ACM SIGKDD international conference on Knowledge discovery and data mining},
  pages={61--69},
  year={2008}
}

@article{bunea2015sample,
  title={On the sample covariance matrix estimator of reduced effective rank population matrices, with applications to fPCA},
  author={Bunea, Florentina and Xiao, Luo},
  year={2015}
}

@article{bauer2000feature,
  title={Feature screening using signal-to-noise ratios},
  author={Bauer Jr, Kenneth W and Alsing, Stephen G and Greene, Kelly A},
  journal={Neurocomputing},
  volume={31},
  number={1-4},
  pages={29--44},
  year={2000},
  publisher={Elsevier}
}

@article{cai2010optimal,
  title={Optimal rates of convergence for covariance matrix estimation},
  author={Cai, T Tony and Zhang, Cun-Hui and Zhou, Harrison H},
  year={2010}
}

@techreport{cifar10,
  title={CIFAR-10 (Canadian Institute for Advanced Research)},
  author={Alex Krizhevsky and Vinod Nair and Geoffrey Hinton},
  year={2009},
  institution={University of Toronto},
  url={https://www.cs.toronto.edu/~kriz/cifar.html}
}

@techreport{cifar100,
  title={CIFAR-100 (Canadian Institute for Advanced Research)},
  author={Alex Krizhevsky and Vinod Nair and Geoffrey Hinton},
  year={2009},
  institution={University of Toronto},
  url={https://www.cs.toronto.edu/~kriz/cifar.html}
}

@article{dhal2022comprehensive,
  title={A comprehensive survey on feature selection in the various fields of machine learning},
  author={Dhal, P. and Azad, C.},
  journal={Applied Intelligence},
  volume={52},
  number={4},
  pages={4543--4581},
  year={2022},
  publisher={Springer}
}

@inproceedings{dhar2019learning,
  title={Learning without memorizing},
  author={Dhar, Prithviraj and Singh, Rajat Vikram and Peng, Kuan-Chuan and Wu, Ziyan and Chellappa, Rama},
  booktitle={Proceedings of the IEEE/CVF conference on computer vision and pattern recognition},
  pages={5138--5146},
  year={2019}
}

@inproceedings{douillard2020podnet,
  title={Podnet: Pooled outputs distillation for small-tasks incremental learning},
  author={Douillard, Arthur and Cord, Matthieu and Ollion, Charles and Robert, Thomas and Valle, Eduardo},
  booktitle={European Conference on Computer Vision},
  pages={86--102},
  year={2020},
  organization={Springer}
}

@InProceedings{Gao_2025_CVPR,
    author    = {Gao, Zijian and Jia, Wangwang and Zhang, Xingxing and Zhou, Dulan and Xu, Kele and Dawei, Feng and Dou, Yong and Mao, Xinjun and Wang, Huaimin},
    title     = {Knowledge Memorization and Rumination for Pre-trained Model-based Class-Incremental Learning},
    booktitle = {Proceedings of the IEEE/CVF Conference on Computer Vision and Pattern Recognition (CVPR)},
    month     = {June},
    year      = {2025},
    pages     = {20523-20533}
}

@techreport{ghahramani1996algorithm,
  title={The EM algorithm for mixtures of factor analyzers},
  author={Ghahramani, Zoubin and Hinton, Geoffrey E and others},
  year={1996},
  institution={CRG-TR-96-1, University of Toronto}
}

@article{girshick1936principal,
  title={Principal components},
  author={Girshick, MA},
  journal={JASA},
  volume={31},
  number={195},
  pages={519--528},
  year={1936},
  publisher={Taylor \& Francis}
}

@article{girshick1939sampling,
  title={On the sampling theory of roots of determinantal equations},
  author={Girshick, MA},
  journal={Ann. Math. Stat.},
  volume={10},
  number={3},
  pages={203--224},
  year={1939},
  publisher={JSTOR}
}

@inproceedings{guo2021network,
  title={Network pruning via annealing and direct sparsity control},
  author={Guo, Yangzi and She, Yiyuan and Barbu, Adrian},
  booktitle={IJCNN},
  pages={1--8},
  year={2021}
}

@article{hamadache2017principal,
  title={Principal component analysis based signal-to-noise ratio improvement for inchoate faulty signals: Application to ball bearing fault detection},
  author={Hamadache, Moussa and Lee, Dongik},
  journal={International Journal of Control, Automation and Systems},
  volume={15},
  number={2},
  pages={506--517},
  year={2017},
  publisher={Springer}
}

@article{hamraz2023feature,
  title={Feature selection for high dimensional microarray gene expression data via weighted signal to noise ratio},
  author={Hamraz, Muhammad and Ali, Amjad and Mashwani, Wali Khan and Aldahmani, Saeed and Khan, Zardad},
  journal={PloS one},
  volume={18},
  number={4},
  pages={e0284619},
  year={2023},
  publisher={Public Library of Science San Francisco, CA USA}
}

@article{hinton2015distilling,
  title={Distilling the knowledge in a neural network},
  author={Hinton, Geoffrey and Vinyals, Oriol and Dean, Jeff},
  journal={arXiv preprint arXiv:1503.02531},
  year={2015}
}

@inproceedings{hou2019learning,
  title={Learning a unified classifier incrementally via rebalancing},
  author={Hou, Saihui and Pan, Xinyu and Loy, Chen Change and Wang, Zilei and Lin, Dahua},
  booktitle={Proceedings of the IEEE/CVF conference on computer vision and pattern recognition},
  pages={831--839},
  year={2019}
}

@inproceedings{huang2003comparative,
  title={A comparative study of feature selection methods for probabilistic neural networks in cancer classification},
  author={Huang, Chenn-Jung and Liao, Wei-Chen},
  booktitle={Int. Conf. on Tools with AI},
  pages={451--458},
  year={2003},
  organization={IEEE}
}

@article{hussain2024optimal,
  title={Optimal features selection in the high dimensional data based on robust technique: Application to different health database},
  author={Hussain, Ibrar and Qureshi, Moiz and Ismail, Muhammad and Iftikhar, Hasnain and Zywio{\l}ek, Justyna and L{\'o}pez-Gonzales, Javier Linkolk},
  journal={Heliyon},
  volume={10},
  number={17},
  year={2024},
  publisher={Elsevier}
}

@article{ILSVRC15,
Author = {Olga Russakovsky and Jia Deng and Hao Su and Jonathan Krause and Sanjeev Satheesh and Sean Ma and Zhiheng Huang and Andrej Karpathy and Aditya Khosla and Michael Bernstein and Alexander C. Berg and Fei-Fei Li},
Title = {{Img.Net Large Scale Visual Recognition Challenge}},
Year = {2015},
journal   = {IJCV},
doi = {10.1007/s11263-015-0816-y},
volume={115},
number={3},
pages={211-252}}

@article{johnstone2001distribution,
  title={On the distribution of the largest eigenvalue in principal components analysis},
  author={Johnstone, Iain M},
  journal={Ann. Stat.},
  volume={29},
  number={2},
  pages={295--327},
  year={2001},
  publisher={Institute of Mathematical Statistics}
}

@inproceedings{kansabanik2025feature,
  title={Feature Selection for Latent Factor Models},
  author={Kansabanik, Rittwika and Barbu, Adrian},
  booktitle={Proceedings of the Computer Vision and Pattern Recognition Conference},
  pages={30742--30751},
  year={2025}
}

@inproceedings{khalid2014survey,
  title={A survey of feature selection and feature extraction techniques in machine learning},
  author={Khalid, Samina and Khalil, Tehmina and Nasreen, Shamila},
  booktitle={Science and information conference},
  pages={372--378},
  year={2014},
  organization={IEEE}
}

@article{koltchinskii2017concentration,
  title={Concentration inequalities and moment bounds for sample covariance operators},
  author={Koltchinskii, V. and Lounici, K.},
  journal={Bernoulli},
  volume={23},
  number={1},
  pages={110--133},
  year={2017},
  publisher={Bernoulli Society for Mathematical Statistics and Probability}
}

@article{koltchinskii2017normal,
  title={Normal approximation and concentration of spectral projectors of sample covariance},
  author={Koltchinskii, Vladimir and Lounici, Karim},
  year={2017}
}

@article{kingma2014adam,
  title={Adam: A method for stochastic optimization},
  author={Kingma, Diederik P and Ba, Jimmy},
  journal={arXiv preprint arXiv:1412.6980},
  year={2014}
}

@article{lawley1962factor,
  title={Factor analysis as a statistical method},
  author={Lawley, Derrick Norman and Maxwell, Albert Ernest},
  journal={Journal of the Royal Statistical Society. Series D (The Statistician)},
  volume={12},
  number={3},
  pages={209--229},
  year={1962},
  publisher={JSTOR}
}

@article{lee2010convergence,
  title={Convergence and prediction of principal component scores in high-dimensional settings},
  author={Lee, Seunggeun and Zou, Fei and Wright, Fred A},
  journal={Ann. Stat.},
  volume={38},
  number={6},
  pages={3605},
  year={2010}
}

@article{lim2023low,
  title={Low-rank learning for feature selection in multi-label classification},
  author={Lim, Hyunki},
  journal={Pattern Recognition Letters},
  volume={172},
  pages={106--112},
  year={2023},
  publisher={Elsevier}
}

@article{mishra2011feature,
  title={Feature selection for cancer classification: a signal-to-noise ratio approach},
  author={Mishra, Debahuti and Sahu, Barnali},
  journal={Int. J. of Scientific \& Eng. Research},
  volume={2},
  number={4},
  pages={1--7},
  year={2011}
}

@article{nadler2008finite,
  title={Finite Sample Approximation Results for Principal Component Analysis: A Matrix Perturbation Approach},
  author={Nadler, Boaz},
  journal={Ann. Stat.},
  pages={2791--2817},
  year={2008}
}

@inproceedings{nakata2022revisiting,
  title={Revisiting a knn-based image classification system with high-capacity storage},
  author={Nakata, Kengo and Ng, Youyang and Miyashita, Daisuke and Maki, Asuka and Lin, Yu-Chieh and Deguchi, Jun},
  booktitle={European conference on computer vision},
  pages={457--474},
  year={2022},
  organization={Springer}
}

@inproceedings{niu2010facial,
  title={Facial expression recognition based on weighted principal component analysis and support vector machines},
  author={Niu, Zhiguo and Qiu, Xuehong},
  booktitle={Int. Conf. on Advanced Computer Theory and Engineering},
  volume={3},
  pages={V3--174},
  year={2010},
  organization={IEEE}
}

@article{oh2024minimum,
  title={Minimum description feature selection for complexity reduction in machine learning-based wireless positioning},
  author={Oh, Myeung Suk and Das, Anindya Bijoy and Kim, Taejoon and Love, David J and Brinton, Christopher G},
  journal={IEEE Journal on Selected Areas in Communications},
  volume={42},
  number={9},
  pages={2585--2600},
  year={2024},
  publisher={IEEE}
}

@article{paul2007asymptotics,
  title={Asymptotics of sample eigenstructure for a large dimensional spiked covariance model},
  author={Paul, Debashis},
  journal={Statistica Sinica},
  pages={1617--1642},
  year={2007},
  publisher={JSTOR}
}

@inproceedings{pushpalatha2023cardio,
  title={Cardio Vascular Disease Prediction Based on PCA-ReliefF Hybrid Feature Selection Method with SVM},
  author={Pushpalatha, L and Durga, R},
  booktitle={International Conference on Advancements in Smart Computing and Information Security},
  pages={40--54},
  year={2023},
  organization={Springer}
}

@misc{radford2021clip,
  author = {Alec Radford and Jong Wook Kim and Chris Hallacy and Aditya Ramesh and Gabriel Goh and Sandhini Agarwal and Girish Sastry and Amanda Askell and Pamela Mishkin and Jack Clark and Gretchen Krueger and Ilya Sutskever},
  title = {{CLIP}: Connecting Text and Images},
  year = {2021},
  howpublished = {\url{https://github.com/openai/CLIP}},
  note = {Accessed: <2024-05-05>}
}

@article{rahmat2024supervised,
  title={Supervised feature selection using principal component analysis},
  author={Rahmat, Fariq and Zulkafli, Zed and Ishak, Asnor Juraiza and Abdul Rahman, Ribhan Zafira and Stercke, Simon De and Buytaert, Wouter and Tahir, Wardah and Ab Rahman, Jamalludin and Ibrahim, Salwa and Ismail, Muhamad},
  journal={Knowledge and Information Systems},
  volume={66},
  number={3},
  pages={1955--1995},
  year={2024},
  publisher={Springer}
}

@inproceedings{rebuffi2017icarl,
  title={icarl: Incremental classifier and representation learning},
  author={Rebuffi, Sylvestre-Alvise and Kolesnikov, Alexander and Sperl, Georg and Lampert, Christoph H},
  booktitle={Proceedings of the IEEE conference on Computer Vision and Pattern Recognition},
  pages={2001--2010},
  year={2017}
}

@article{rubin1982algorithms,
  title={EM algorithms for ML factor analysis},
  author={Rubin, Donald B and Thayer, Dorothy T},
  journal={Psychometrika},
  volume={47},
  pages={69--76},
  year={1982},
  publisher={Springer}
}

@article{salas2010feature,
  title={Feature selection using factor analysis for Alzheimer's diagnosis using PET images},
  author={Salas-Gonzalez, Diego and G{\'o}rriz, JM and Ram{\'\i}rez, Javier and Ill{\'a}n, IA and L{\'o}pez, Miriam and Segovia, F and Chaves, R and Padilla, P and Puntonet, CG and Alzheimer's Disease Neuroimage Initiative},
  journal={Medical physics},
  volume={37},
  number={11},
  pages={6084--6095},
  year={2010},
  publisher={Wiley Online Library}
}

@article{sahu2017feature,
  title={Feature selection model based on clustering and ranking in pipeline for microarray data},
  author={Sahu, Barnali and Dehuri, Satchidananda and Jagadev, Alok Kumar},
  journal={Informatics in Medicine Unlocked},
  volume={9},
  pages={107--122},
  year={2017},
  publisher={Elsevier}
}

@article{simeoni2025dinov3,
  title={Dinov3},
  author={Sim{\'e}oni, Oriane and Vo, Huy V and Seitzer, Maximilian and Baldassarre, Federico and Oquab, Maxime and Jose, Cijo and Khalidov, Vasil and Szafraniec, Marc and Yi, Seungeun and Ramamonjisoa, Micha{\"e}l and others},
  journal={arXiv preprint arXiv:2508.10104},
  year={2025}
}

@article{sharifzadeh2017sparse,
  title={Sparse supervised principal component analysis (SSPCA) for dimension reduction and variable selection},
  author={Sharifzadeh, Sara and Ghodsi, Ali and Clemmensen, Line H and Ersb{\o}ll, Bjarne K},
  journal={Engineering Applications of Artificial Intelligence},
  volume={65},
  pages={168--177},
  year={2017},
  publisher={Elsevier}
}

@article{she2012iterative,
  title={An iterative algorithm for fitting nonconvex penalized generalized linear models with grouped predictors},
  author={She, Yiyuan},
  journal={Computational Statistics \& Data Analysis},
  volume={56},
  number={10},
  pages={2976--2990},
  year={2012},
  publisher={Elsevier}
}

@article{she2009thresholding,
  title={Thresholding-based iterative selection procedures for model selection and shrinkage},
  author={She, Yiyuan},
  journal={Elec. J. of Stat.},
  volume={3},
  pages={384--415},
  year={2009}
}

@article{tipping1999probabilistic,
  title={Probabilistic principal component analysis},
  author={Tipping, Michael E and Bishop, Christopher M},
  journal={JRSS B},
  volume={61},
  number={3},
  pages={611--622},
  year={1999},
  publisher={Oxford University Press}
}

@article{townes2019feature,
  title={Feature selection and dimension reduction for single-cell RNA-Seq based on a multinomial model},
  author={Townes, F William and Hicks, Stephanie C and Aryee, Martin J and Irizarry, Rafael A},
  journal={Genome biology},
  volume={20},
  number={1},
  pages={295},
  year={2019},
  publisher={Springer}
}

@article{uuguz2011two,
  title={A two-stage feature selection method for text categorization by using information gain, principal component analysis and genetic algorithm},
  author={U{\u{g}}uz, Harun},
  journal={Knowledge-Based Systems},
  volume={24},
  number={7},
  pages={1024--1032},
  year={2011},
  publisher={Elsevier}
}

@book{wainwright2019high,
  title={High-dimensional statistics: A non-asymptotic viewpoint},
  author={Wainwright, Martin J},
  volume={48},
  year={2019},
  publisher={Cambridge university press}
}

@INPROCEEDINGS{wang2023scalable,
  author={Wang, Boshi and Barbu, Adrian},
  booktitle={IEEE Int. Conf. on Big Data}, 
  title={Scalable Learning with Incremental Probabilistic PCA}, 
  year={2022},
  volume={},
  number={},
  pages={5615-5622},
  doi={10.1109/BigData55660.2022.10020330}}

@article{waternaux1976asymptotic,
  title={Asymptotic distribution of the sample roots for a nonnormal population},
  author={Waternaux, Christine M},
  journal={Biometrika},
  volume={63},
  number={3},
  pages={639--645},
  year={1976},
  publisher={Oxford University Press}
}

@article{wu1983convergence,
  author  = {Wu, C. F. J.},
  title   = {On the Convergence Properties of the {EM} Algorithm},
  journal = {The Annals of Statistics},
  volume  = {11},
  number  = {1},
  pages   = {95--103},
  year    = {1983}
}

@inproceedings{wu2019large,
  title={Large scale incremental learning},
  author={Wu, Yue and Chen, Yinpeng and Wang, Lijuan and Ye, Yuancheng and Liu, Zicheng and Guo, Yandong and Fu, Yun},
  booktitle={Proceedings of the IEEE/CVF conference on computer vision and pattern recognition},
  pages={374--382},
  year={2019}
}

@inproceedings{yu2006supervised,
  title={Supervised probabilistic principal component analysis},
  author={Yu, Shipeng and Yu, Kai and Tresp, Volker and Kriegel, Hans-Peter and Wu, Mingrui},
  booktitle={SIGKDD},
  pages={464--473},
  year={2006}
}

@article{zagoruyko2016paying,
  title={Paying more attention to attention: Improving the performance of convolutional neural networks via attention transfer},
  author={Zagoruyko, Sergey and Komodakis, Nikos},
  journal={arXiv preprint arXiv:1612.03928},
  year={2016}
}

@article{zhang2022heteroskedastic,
  title={Heteroskedastic PCA: Algorithm, optimality, and applications},
  author={Zhang, Anru R and Cai, T Tony and Wu, Yihong},
  journal={Ann. Stat.},
  volume={50},
  number={1},
  pages={53--80},
  year={2022},
  publisher={Institute of Mathematical Statistics}
}

@article{zhivotovskiy2024dimension,
  title={Dimension-free bounds for sums of independent matrices and simple tensors via the variational principle},
  author={Zhivotovskiy, Nikita},
  journal={Electronic Journal of Probability},
  volume={29},
  pages={1--28},
  year={2024},
  publisher={The Institute of Mathematical Statistics and the Bernoulli Society}
}

@article{zhou2014face,
  title={Face recognition based on PCA and logistic regression analysis},
  author={Zhou, Changjun and Wang, Lan and Zhang, Qiang and Wei, Xiaopeng},
  journal={Optik},
  volume={125},
  number={20},
  pages={5916--5919},
  year={2014},
  publisher={Elsevier}
}

@article{zhou2025revisiting,
  title={Revisiting class-incremental learning with pre-trained models: Generalizability and adaptivity are all you need},
  author={Zhou, Da-Wei and Cai, Zi-Wen and Ye, Han-Jia and Zhan, De-Chuan and Liu, Ziwei},
  journal={International Journal of Computer Vision},
  volume={133},
  number={3},
  pages={1012--1032},
  year={2025},
  publisher={Springer}
}

@article{li2018learning,
  author = {Li, Z. and Hoiem, D.},
  title = {Learning without Forgetting},
  journal = {IEEE TPAMI},
  year = {2018}
}

@article{wang2022learning,
  author = {Wang, Z. and Zhang, Z. and Luo, Y. and Ebrahimi, S. and Sun, Y. and Liu, H.},
  title = {Learning to Prompt for Continual Learning},
  journal = {CVPR},
  year = {2022}
}

@article{wang2022dualprompt,
  author = {Wang, Z. and Zhang, Z. and Luo, Y. and Ebrahimi, S. and Sun, Y. and Liu, H.},
  title = {DualPrompt: Complementary Prompting for Rehearsal-free Continual Learning},
  journal = {ECCV},
  year = {2022}
}

@article{ye2022acil,
  author = {Ye, H.-J. and Xie, C.-X. and Liu, H. and Zhan, D.-C.},
  title = {Adversarial Consistency for Incremental Learning},
  journal = {NeurIPS},
  year = {2022}
}

@article{smith2023coda,
  author = {Smith, J. S. and Karlinsky, M. and Goyal, V. and Feris, R. and Kirillov, A. and Elhoseiny, M. and Ghanem, B.},
  title = {CODA-Prompt: COntinual Decomposed Attention-based Prompting for Rehearsal-Free Continual Learning},
  journal = {CVPR},
  year = {2023}
}

@article{liu2023lae,
  author = {Liu, Y. and Wang, L. and Liu, H. and Ye, H.-J. and Zhan, D.-C.},
  title = {Learning from Ambiguous Examples for Incremental Learning},
  journal = {ICCV},
  year = {2023}
}

@article{chen2024dsal,
  author = {Chen, Y. and Wang, L. and Liu, H. and Zhan, D.-C.},
  title = {Dynamic Subspace Adaptation for Continual Learning},
  journal = {AAAI},
  year = {2024}
}

@article{li2024ease,
  author = {Li, Z. and Liu, H. and Ye, H.-J. and Zhan, D.-C.},
  title = {EASE: Efficient and Stable Continual Learning with Adapters},
  journal = {CVPR},
  year = {2024}
}
}

\clearpage
\section{Supplementary}
The theoretical investigation is motivated explicitly by our recent work in \cite{kansabanik2025feature}, where we introduced a novel and highly effective method for feature selection in multi-class classification problems. 
The core of that method was the use of the Signal-to-Noise Ratio (SNR), derived from class-specific LFA models, as a criterion for identifying and ranking informative features. 
While the method was vindicated by strong empirical performance, it was also supported by rigorous theoretical analysis. 
The previous paper established the asymptotic consistency of the SNR estimator, demonstrating that under certain assumptions, the estimated SNR converges in probability to the actual SNR as the number of samples $n$ approaches infinity. 

\begin{theorem}\label{thm:recv}[from \cite{kansabanik2025feature}]
Consider the following assumptions:
\begin{itemize}
\item[(\textbf{A1})] The observations, $(\bx_i,i=1,2,\cdots,n)$ are independently generated from the LFA model \eqref{eq:LFAm} with $\bmu = 0$ and $\beps \overset{i.i.d}{\sim}\N(\bzero,\bPsi)$. 

\item[(\textbf{A2})] Denote $\bGamma = (\bgamma_1,\bgamma_2,\cdots,\bgamma_n)^T$. $\bGamma_{ij}$ are i.i.d random variables with $E(\bGamma_{ij}) = 0, var(\bGamma_{ij}) = 1, E(\bGamma_{ij}^4) < \infty$, for all $(i,j) \in \{1,2,\cdots,n\}\times \{1,2,\cdots,r\}$. 

\item[(\textbf{A3})] There are $m$ true features with indices $S=\{s_1,s_2,\cdots, s_m\}$ and $(d-m)$ noisy features in our data. 
For a positive constant, $\gamma >0$, we have,
\begin{equation*}
\min\{\bSNR_{i}, i \in S\} \ge \max\{\bSNR_{i}, i \not \in S\} + \gamma.
\end{equation*}
\end{itemize}
Let $d$ be fixed, and let $n \to \infty$.
\begin{itemize}
\item[(\textbf{C1})] Under assumptions (\textbf{A1},\textbf{A2}), if $\bPsi = \sigma^{\ast 2} \bI_d$, then 

\[
    \hat{\bSNR}^{PPCA}_i \overset{p}{\to} \bSNR_i,
    \]
    for all $i \in \{1, 2, \dots, d\}$.
    
\item[(\textbf{C2})] Under the assumption (\textbf{A1}), if $\bPsi = \diag(\sigma_1^{ 2}, \sigma_2^{ 2}, \dots, \sigma_d^{ 2})$ and $\bgamma_i \overset{i.i.d}{\sim} \N(\bzero, \bI_r)$ then
\vspace{-2mm}
\[
    \hat{\bSNR}^{LFA}_i \overset{p}{\to} \bSNR_i,
\vspace{-2mm}
    \]
    for all $i \in \{1, 2, \dots, d\}$.
\end{itemize}
Furthermore, under the assumption (\textbf{A3}), the probability that the $m$ features with the highest SNRs are the true features converges to 1 as $n \to \infty$ for both (\textbf{C1}) and (\textbf{C2}).
\end{theorem}

Here, $\hat{\bSNR}^{PPCA}$ and $\hat{\bSNR}^{LFA}$ denote the estimated SNRs from PPCA and LFA, respectively.

This result, in turn, provided an asymptotic guarantee of true feature recovery. 
It demonstrates that the probability of selecting the correct feature set converges to one in the limit. 
Such asymptotic guarantees are foundational and motivate us to characterize the estimator's performance in the practical, non-limiting regime. 
This paper is therefore dedicated to building upon that foundational work by providing the necessary non-asymptotic analysis.

Our analysis begins with bounding the deviation of the sample covariance matrix
$\hat{\bSigma} = \frac{1}{n}\sum_{i=1}^n \bx_i \bx_i^T$
from its population counterpart $\bSigma$. 
This deviation directly governs the accuracy of estimated signal and noise variances in PPCA and LFA.

We will use two complementary results: a \textit{global bound} on the operator norm $\|\hat{\bSigma} - \bSigma\|_{op}$, which controls the spectral deviation of all eigenvalues collectively, and a \textit{local bound} on the diagonal elements $|(\hat{\bSigma} - \bSigma)_{ii}|$, which quantifies feature-wise deviations. 
The global bound is due to \cite{zhivotovskiy2024dimension}.

\begin{theorem}[from \cite{zhivotovskiy2024dimension}][Covariance Matrix Concentration]
\label{thm:zhivotovskiy}
Assume that $M_1, \dots, M_n$ are independent realizations of a $d \times d$ positive semi-definite symmetric random matrix $M$ with mean $\bE[M] = \bSigma$. 
Let $M$ satisfy for some $\kappa \geq 1$,
\begin{equation}
    \|\bx^\top M \bx\|_{\psi_1} \leq \kappa^2 \bx^\top \bSigma \bx, \quad \text{for all } \bx \in \RR^d,
\end{equation}
where $\|\cdot\|_{\psi_1}$ denotes the sub-exponential norm, defined as
$\|\bY\|_{\psi_\alpha} = \inf\{c > 0 : \mathbf{E}[\exp(|\bY|^\alpha / c^\alpha)] \le 2\}$.
Then, for any $t > 0$, with probability at least $1 - \exp(-t)$, it holds that
\begin{equation}
    \|\frac{1}{n} \sum_{i=1}^{n} M_i - \bSigma\|_{op} \leq 20\kappa^2 \|\bSigma\|_{op} \sqrt{\frac{4r(\bSigma) + t}{n}},
\end{equation}
whenever $n \geq 4r(\bSigma) + t$, where $r(\bSigma) = \tr(\bSigma) / \|\bSigma\|_{op}$ is the effective rank of $\bSigma$.
\end{theorem}
For many latent variable models, the observed data vectors $\bx_i$ are assumed to be zero-mean sub-Gaussian random vectors. 
In this case, we can set $M_i = \bx_i \bx_i^T$. 
For a centered Gaussian vector $\bx_i \sim \mathcal{N}(0, \bSigma)$, the condition of Theorem \ref{thm:zhivotovskiy} is met with $\kappa^2 = 8/3$. 
Considerable research \cite{cai2010optimal,adamczak2011sharp,bunea2015sample,koltchinskii2017normal, koltchinskii2017concentration} has focused on the problem of deriving non-asymptotic bounds for the operator norm error $\|\hat{\bSigma} - \bSigma\|_{op}$. 
Theorem~\ref{thm:zhivotovskiy} is a special case of the deviation bound derived by \cite{koltchinskii2017concentration}, which provides a dimension(d)-free upper bound, ensuring the stability of eigenvalue-based estimators such as PPCA loadings. 
It implies that accurate signal recovery is possible whenever $n \gg r(\Sigma)$, even if $d \gg n$. 
The effective rank term $r(\Sigma)$ naturally emerges as a dimension-corrected complexity measure.

While the global bound provides a powerful worst-case guarantee on the overall deviation of the matrix in any direction, we also employ a tighter, more refined bound on the error of the individual diagonal elements, $|\hat{\bSigma}_{ii} - \bSigma_{ii}|$. 
If the feature variances are highly heterogeneous, the global bound can yield a loose bound for low-variance features. 
A tighter bound for individual diagonal elements, $|(\hat{\bSigma} - \bSigma)_{ii}|$, can be obtained by applying a scalar concentration inequality directly, as stated in Theorem \ref{thm:scalar_bernstein}.

\subsection{Proofs}

\begin{proof} of Theorem \ref{thm:scalar_bernstein}.
The term to be bounded is the average of $n$ i.i.d. zero-mean random variables:
\begin{equation}
    \hat{\boldsymbol{\Sigma}}_{ii} - \boldsymbol{\Sigma}_{ii} = \frac{1}{n} \sum_{k=1}^n \left( x_{ki}^2 - \mathbb{E}[x_{ki}^2] \right).
\end{equation}
Let $Z_k = x_{ki}^2 - \mathbb{E}[x_{ki}^2]$. From the LFA model, $x_{ki}$ is a zero-mean Gaussian variable with variance $\boldsymbol{\Sigma}_{ii} = (\boldsymbol{W}\boldsymbol{W}^T)_{ii} + \sigma_i^2$. Thus, $Z_k$ is a centered, scaled chi-squared random variable ($\boldsymbol{\Sigma}_{ii}(\chi_1^2 - 1)$), which is sub-exponential.

To apply Bernstein's inequality\cite{wainwright2019high}, we identify its parameters. The variance parameter $v$ is:
\begin{equation}
    v = \text{Var}(Z_k) = \text{Var}(x_{ki}^2) = 2 \boldsymbol{\Sigma}_{ii}^2.
\end{equation}
The sub-exponential scale parameter $b$ is proportional to the variance of the underlying Gaussian, so $b = C_b \boldsymbol{\Sigma}_{ii}$ for some universal constant $C_b$. For a $\chi_1^2$ variable, a standard version of Bernstein's inequality uses parameters equivalent to $v=2\boldsymbol{\Sigma}_{ii}^2$ and $b=4\boldsymbol{\Sigma}_{ii}$.

Bernstein's inequality for the average provides the following bound on the tail probability:
\begin{equation}
    \mathbb{P}\left( \left| \frac{1}{n} \sum_{k=1}^n Z_k \right| \ge \epsilon \right) \le 2 \exp\left( -c \cdot n \cdot \min\left( \frac{\epsilon^2}{2\boldsymbol{\Sigma}_{ii}^2}, \frac{\epsilon}{4\boldsymbol{\Sigma}_{ii}} \right) \right),
\end{equation}
where $c$ is a universal constant.
Setting the right-hand side to our desired failure probability $\delta/2$ and solving for $\epsilon$ gives:
\begin{equation}
    \epsilon \ge \max\left( \sqrt{\frac{2\boldsymbol{\Sigma}_{ii}^2 \ln(4/\delta)}{cn}}, \frac{4\boldsymbol{\Sigma}_{ii} \ln(4/\delta)}{cn} \right).
\end{equation}
This can be rewritten as:
\begin{equation} \label{eq:max_form_bound}
    \epsilon \ge \boldsymbol{\Sigma}_{ii} \cdot \max\left( \sqrt{\frac{2 \ln(4/\delta)}{cn}}, \frac{4 \ln(4/\delta)}{cn} \right).
\end{equation}
The max operator selects between the sub-Gaussian-like term (with $\sqrt{1/n}$) and the sub-exponential-like term (with $1/n$). The first term is larger if and only if:
\begin{equation}
    \sqrt{\frac{2 \ln(4/\delta)}{cn}} > \frac{4 \ln(4/\delta)}{cn}.
\end{equation}
Squaring both sides (which are positive) yields:
\begin{equation}
    \frac{2 \ln(4/\delta)}{cn} > \frac{16 (\ln(4/\delta))^2}{c^2 n^2}.
\end{equation}
Assuming $\ln(4/\delta) > 0$, we can simplify by multiplying by $c^2 n^2$ and dividing by $c n \ln(4/\delta)$:
\begin{equation}
    2c > \frac{16 \ln(4/\delta)}{n}.
\end{equation}
Rearranging gives the constraint on $n$:
\begin{equation} \label{eq:n_constraint}
    n > \frac{8}{c} \ln(4/\delta).
\end{equation}

Under the constraint in Eq. \eqref{eq:n_constraint}, the first term in Eq. \eqref{eq:max_form_bound} dominates. The bound on the error simplifies to:
\begin{equation}
    \left|\hat{\boldsymbol{\Sigma}}_{ii} - \boldsymbol{\Sigma}_{ii}\right| \le \boldsymbol{\Sigma}_{ii} \sqrt{\frac{2 \ln(4/\delta)}{cn}}.
\end{equation}
\end{proof}
To analyze the error in the estimated signal variance for a single feature, we first establish a key identity.

\begin{lem}[\textbf{PPCA Signal Decomposition}] \label{lem:ppcasdec}
If the assumptions $(\textbf{A1-A2})$ hold, the $i$-th diagonal element of the PPCA signal estimate $(\widehat{\bW\bW^T})_{ii}$ can be written as:
\begin{equation}\label{eq:ppca_error_decomp}
\begin{split}
&(\widehat{\bW\bW^T})_{ii} = \hat{\bSigma}_{ii} - \hat{\sigma}^2 - \Delta_i,\\
&\abs{ (\bW\bW^T)_{ii} - (\widehat{\bW\bW^T})_{ii} } \le \abs{(\hat{\bSigma}- \bSigma)_{ii}} + \abs{\hat{\sigma}^2 - \sigma^2 +\Delta_i},
\end{split}
\end{equation}
where $(\hat{\bW},\hat{\sigma}^2)$ is defined in \eqref{eq:cest}, $l_j$ is the $j^{th}$ eigenvalue and $\bu_j$ is the $j^{th}$ eigen vector of $\hat{\bSigma}$, and $\Delta_i = \sum_{j=r+1}^d (l_j - \hat{\sigma}^2) (\bu_j)_i^2$ is a remainder term.
\end{lem}
\begin{proof}
The $i$-th diagonal element of $\hat{\bSigma}$ is $\hat{\bSigma}_{ii} = \sum_{j=1}^d l_j (\bu_j)_i^2$. The eigenvectors form an orthonormal basis, so $\sum_{j=1}^d (\bu_j)_i^2 = 1$.
The estimated signal variance is $(\widehat{\bW\bW^T})_{ii} = \sum_{j=1}^r (l_j - \hat{\sigma}^2) (\bu_j)_i^2$.
\begin{equation}\label{eq:PPCA-sig-dicomp}
    \begin{split}
        (\widehat{\bW\bW^T})_{ii} &= \sum_{j=1}^r l_j (\bu_j)_i^2 - \hat{\sigma}^2 \sum_{j=1}^r (\bu_j)_i^2 \\
&= \left(\hat{\bSigma}_{ii} - \sum_{j=r+1}^d l_j (\bu_j)_i^2\right) - \hat{\sigma}^2 \left(1 - \sum_{j=r+1}^d (\bu_j)_i^2\right) \\
&= \hat{\bSigma}_{ii} - \hat{\sigma}^2 - \left(\sum_{j=r+1}^d l_j (\bu_j)_i^2 - \hat{\sigma}^2 \sum_{j=r+1}^d (\bu_j)_i^2\right) \\
&= \hat{\bSigma}_{ii} - \hat{\sigma}^2 - \sum_{j=r+1}^d (l_j - \hat{\sigma}^2) (\bu_j)_i^2.
    \end{split}
\end{equation}

We have:
$\bSigma = \bW\bW^T + \sigma^2 \bI_d$

Then, the difference is:
\begin{equation}
    \begin{split}
        & \quad | (\mathbf{W}\mathbf{W}^T)_{ii} - (\hat{\mathbf{W}}\hat{\mathbf{W}}^T)_{ii} | \\
        &= | (\bSigma_{ii} - \sigma^2) - (\hat{\bSigma}_{ii} - \hat{\sigma}^2 - \Delta_i) | \\
        &= | (\bSigma - \hat{\bSigma})_{ii} - (\sigma^2 - \hat{\sigma}^2) + \Delta_i | \\
        &\le | (\bSigma - \hat{\bSigma})_{ii} | + | \sigma^2 - \hat{\sigma}^2  + \Delta_i |
    \end{split}
\end{equation}
\end{proof}

\begin{proof}[Theorem \ref{thm:ppca-signal}]
From the equation \eqref{eq:PPCA-sig-dicomp}, we get
$(\bW\bW^T)_{ii} = \bSigma_{ii} - \sigma^2$, as $\sigma^2 > 0$, therefore $(\bW\bW^T)_{ii} < \bSigma_{ii}$.
On the other hand, 
\[
(\hat{\bW}\hat{\bW}^T)_{ii} = \hat{\bSigma}_{ii} - \left[ \sum_{j = r+1}^d l_j(\bu_j)^2_i + \hat{\sigma}^2 \sum_{j = 1}^r (\bu_j)_i^2 \right] 
\]
as $l_j \ge 0 \text{ for all } j \geq 1$ since $\hat{\bSigma}$ is psd and $\hat{\sigma}^2>0$, thus $(\hat{\bW}\hat{\bW}^T)_{ii} < \hat{\bSigma}_{ii}$.

Since $\bW\bW^T$ and $\hat{\bW}\hat{\bW}^T$ are positive semidefinite, their diagonal entries are nonnegative, so $0 \le (\bW\bW^T)_{ii} \le \bSigma_{ii}$ and $0 \le (\hat{\bW}\hat{\bW}^T)_{ii} \le \hat{\bSigma}_{ii}$. Hence
\begin{align}\label{ppca-ini}
\big|(\hat{\bW}\hat{\bW}^T)_{ii} - (\bW\bW^T)_{ii}\big|
&\leq \max\{\hat{\bSigma}_{ii},\, \bSigma_{ii}\} \notag\\
&\leq \big|\hat{\bSigma}_{ii} - \bSigma_{ii}\big| + \bSigma_{ii}.
\end{align}

On the other hand, the error decomposition established in Eq. \eqref{eq:ppca_error_decomp} from Lemma \ref{lem:ppcasdec} above:
$$
\left| (\mathbf{W}\mathbf{W}^T)_{ii} - (\widehat{\mathbf{W}\mathbf{W}^T})_{ii} \right| \le \left|(\hat{\mathbf{\Sigma}}- \mathbf{\Sigma})_{ii}\right| + \left|\hat{\sigma}^2 - \sigma^2 + \Delta_i\right|.
$$
We bound each of the two terms using a union bound, allocating a probability of failure of $\delta/2$ to the first term and $\delta/2$ to the events driven by the operator norm.

\textbf{Term 1: $\left|(\hat{\mathbf{\Sigma}}- \mathbf{\Sigma})_{ii}\right|$}. This term represents the deviation of a single element of the sample covariance matrix. As established in \ref{thm:scalar_bernstein}, the average of the i.i.d. mean-zero sub-exponential variables $Z_k = (\mathbf{x}_k)_i^2 - \mathbf{\Sigma}_{ii}$ is bounded by applying Bernstein's inequality. For $n > \frac{8}{c} \ln(4/\delta)$, with a failure probability of $\delta/2$, the bound is:
    \begin{equation}
        \abs{(\hat{\bSigma} - \bSigma)_{ii}} \le \bSigma_{ii} \sqrt{\frac{2\ln(4/\delta)}{cn}}
    \end{equation}

\textbf{Term 2:} We want to bound $|\hat{\sigma}^2 - \sigma^2 + \Delta_i|$.
    \begin{multline}\label{eq:term2,3-ppca}
    \hat{\sigma}^2 - \sigma^2 + \Delta_i = (\hat{\sigma}^2 - \sigma^2) + \left(\sum_{j=r+1}^d (l_j - \hat{\sigma}^2) (\bu_j)_i^2\right) \\
    =(\hat{\sigma}^2 - \sigma^2) + \left(\sum_{j=r+1}^d (l_j - \sigma^2 - (\hat{\sigma}^2 - \sigma^2)) (\bu_j)_i^2\right) \\
    = (\hat{\sigma}^2 - \sigma^2) + \sum_{j=r+1}^d (l_j - \sigma^2)(\bu_j)_i^2 - (\hat{\sigma}^2 - \sigma^2)\sum_{j=r+1}^d (\bu_j)_i^2 \\
    = (\hat{\sigma}^2 - \sigma^2)\left(1 - \sum_{j=r+1}^d (\bu_j)_i^2\right) + \sum_{j=r+1}^d (l_j - \sigma^2)(\bu_j)_i^2.
    \end{multline}
    Using the identity $1 - \sum_{j=r+1}^d (\bu_j)_i^2 = \sum_{j=1}^r (\bu_j)_i^2$, from \eqref{eq:term2,3-ppca}, we have:
    $$
    \hat{\sigma}^2 - \sigma^2 + \Delta_i= (\hat{\sigma}^2 - \sigma^2)\sum_{j=1}^r (\bu_j)_i^2 + \sum_{j=r+1}^d (l_j - \sigma^2)(\bu_j)_i^2.
    $$

Results like Weyl's\cite{wainwright2019high} inequality relate the eigenvalues $l_j$ of $\hat{\bSigma}$ to the eigenvalues $\lambda_j$ of the true covariance $\bSigma = \bW \bW^T + \sigma^2 \bI_d$. Specifically,
    \begin{equation*}
        \max_{j=1, \dots, d} |l_j - \lambda_j(\Sigma)| \leq \|\hat{\bSigma}_n - \Sigma\|_{op}.
    \end{equation*}
    where $\lambda_1 \ge \dots \ge \lambda_d$ are the eigenvalues of $\bSigma$. For PPCA, $\lambda_j = \sigma^2$ for $j \in {r+1,\dots,d}$. Therefore, Using Weyl's inequality, we get:
    \begin{equation}\label{eq:ppca-error}
    \begin{split}
        |\hat{\sigma}^2 - \sigma^2| &= \left| \frac{1}{d-r} \sum_{j=r+1}^d l_j - \sigma^2 \right| 
        = \left| \frac{1}{d-r} \sum_{j=r+1}^d (l_j - \lambda_j)\right|  \\
        &\leq \frac{1}{d-r} \sum_{j=r+1}^d |l_j - \lambda_j| \quad(\text{since } \lambda_j = \sigma^2 \text{ for } j > r)\\
        &\leq \max_{j=r+1, \dots, d} |l_j - \lambda_j| \leq \|\hat{\bSigma} - \bSigma\|_{op}
    \end{split}
    \end{equation}
     Once again, we apply the triangle inequality and Weyl's inequality\cite{wainwright2019high}:
    \begin{multline}
    |\hat{\sigma}^2 - \sigma^2 + \Delta_i| \le |\hat{\sigma}^2 - \sigma^2|\sum_{j=1}^r (\bu_j)_i^2 + \sum_{j=r+1}^d |l_j - \sigma^2|(\bu_j)_i^2 \\
    \le \|\hat{\mathbf{\Sigma}}- \mathbf{\Sigma}\|_{op} \sum_{j=1}^r (\bu_j)_i^2 + \|\hat{\mathbf{\Sigma}}- \mathbf{\Sigma}\|_{op} \sum_{j=r+1}^d (\bu_j)_i^2 \quad(\text{due to} \eqref{eq:ppca-error})\\
    = \|\hat{\mathbf{\Sigma}}- \mathbf{\Sigma}\|_{op} \left( \sum_{j=1}^r (\bu_j)_i^2 + \sum_{j=r+1}^d (\bu_j)_i^2 \right) \\
    = \|\hat{\mathbf{\Sigma}}- \mathbf{\Sigma}\|_{op} \sum_{j=1}^d (\bu_j)_i^2 = \|\hat{\mathbf{\Sigma}}- \mathbf{\Sigma}\|_{op}.
    \end{multline}

The bound for Term 2 depends on the operator norm bound $\|\hat{\mathbf{\Sigma}}- \mathbf{\Sigma}\|_{op}$. With a probability of at least $1-\delta/2$, we use the dimension-free bound from \cite{zhivotovskiy2024dimension}, which applies to sub-Gaussian vectors. To achieve a failure probability of $\delta' = \delta/2$,we set $\delta/2 = e^{-t}$, which implies $t = \ln(2/\delta)$. Substituting this into the bound from that paper, we have that with probability at least $1-\delta/2$:
\begin{equation}
    \|\hat{\mathbf{\Sigma}} - \mathbf{\Sigma}\|_{op} \le 20 \cdot \frac{8}{3} \cdot \|\mathbf{\Sigma}\|_{op} \cdot \sqrt{\frac{4r(\mathbf{\Sigma}) + \ln(2/\delta)}{n}}.
\end{equation}

The bound for Term 1 also holds with probability at least $1-\delta/2$ whenever $n > \frac{8}{c} \ln(4/\delta)$. 

Therefore with probability at least $1-\delta$:
\begin{multline}\label{ppca-op}
    \left| (\mathbf{W}\mathbf{W}^T)_{ii} - (\widehat{\mathbf{W}\mathbf{W}^T})_{ii} \right| \le \bSigma_{ii} \sqrt{\frac{2\ln(4/\delta)}{cn}} + \\
    \frac{160}{3} \cdot \|\mathbf{\Sigma}\|_{op} \cdot \sqrt{\frac{4r(\mathbf{\Sigma}) + \ln(2/\delta)}{n}}\\
    = \bSigma_{ii} \sqrt{\frac{2\ln(4/\delta)}{cn}} + \epsilon(\delta/2)
\end{multline}

Combining the two inequalities in \eqref{ppca-ini} and,\eqref{ppca-op}, we get:
$\quad |\hat{\bW}\hat{\bW}^T - \bW\bW^T| \leq \bSigma_{ii} \sqrt{\frac{2\ln(4/\delta)}{cn}} + \min\{\epsilon(\delta/2), 2\bSigma_{ii}\}$.
\end{proof}

Proof of the corollary \ref{cor:ppca-recovery}
\begin{proof}
Let $\mathcal{E}_i = \big\{ |\theta_i - \hat{\theta}_i| \le B(\delta) \big\}$ and
$\mathcal{E} = \bigcap_{i=1}^{d}\mathcal{E}_i$.

By Theorem \ref{thm:scalar_bernstein} applied at level $\delta/(2d)$, for each
$i \in \{1,\dots,d\}$,
\begin{equation}
\mathbb{P}\left( \big|\hat{\bSigma}_{ii} - \bSigma_{ii}\big|
> \bSigma_{ii}\sqrt{\frac{2\ln(4d/\delta)}{cn}} \right) \le \frac{\delta}{2d},
\end{equation}
so that, by \eqref{eq:ppca-rec-n-cond} and a union bound over $i$,
\begin{equation}
\mathbb{P}\left( \bigcup_{i=1}^{d}\left\{ \big|\hat{\bSigma}_{ii} - \bSigma_{ii}\big|
> \bSigma_{\max}\sqrt{\frac{2\ln(4d/\delta)}{cn}} \right\} \right) \le \frac{\delta}{2}.
\end{equation}
By Theorem \ref{thm:zhivotovskiy} at level $\delta/2$, valid under
\eqref{eq:ppca-rec-n-cond},
\begin{equation}
\mathbb{P}\Big( \|\hat{\bSigma}-\bSigma\|_{\op} > \epsilon(\delta/2) \Big) \le \frac{\delta}{2},
\end{equation}
and this event does not depend on $i$.
\\
Define the events
\begin{align*}
\mathcal{A} &= \bigcap_{i=1}^{d}\left\{
\big|(\hat{\bSigma}-\bSigma)_{ii}\big| \le
\bSigma_{\max}\sqrt{\frac{2\ln(4d/\delta)}{cn}} \right\},
\\
\mathcal{B} &= \Big\{ \|\hat{\bSigma}-\bSigma\|_{\op} \le \epsilon(\delta/2) \Big\},
\end{align*}
Therefore,
\begin{align*}
    \mathbb{P}(\mathcal{A}\cap\mathcal{B})^c &= \mathbb{P}(\mathcal{A}^c\cup\mathcal{B}^c) \le \mathbb{P}(\mathcal{A}^c) + \mathbb{P}(\mathcal{B}^c) 
    \\&= \mathbb{P}\left( \bigcup_{i=1}^{d}\left\{ \big|\hat{\bSigma}_{ii} - \bSigma_{ii}\big|
    >\bSigma_{\max}\sqrt{\frac{2\ln(4d/\delta)}{cn}} \right\} \right) 
    \\&+ \mathbb{P}\Big( \|\hat{\bSigma}-\bSigma\|_{\op} > \epsilon(\delta/2) \Big) 
    \\&\le \frac{\delta}{2} + \frac{\delta}{2} = \delta
\end{align*}

\begin{equation}
\mathbb{P}(\mathcal{A}\cap\mathcal{B}) \ge 1-\delta.
\label{eq:ppca-rec-AB}
\end{equation}

Fix $i \in \{1,\dots,d\}$ and work on $\mathcal{A}\cap\mathcal{B}$.
By Lemma \ref{lem:ppcasdec},
\begin{equation}
\big|\theta_i - \hat{\theta}_i\big|
\le \big|(\hat{\bSigma}-\bSigma)_{ii}\big| + \big|\hat{\sigma}^2-\sigma^2+\Delta_i\big|,
\label{eq:ppca-rec-decomp}
\end{equation}
and, as established in the proof of Theorem \ref{thm:ppca-signal},
\begin{equation}
\big|\hat{\sigma}^2-\sigma^2+\Delta_i\big| \le \|\hat{\bSigma}-\bSigma\|_{\op} \le \epsilon(\delta/2),
\label{eq:ppca-rec-term2}
\end{equation}
the last step on $\mathcal{B}$. Substituting \eqref{eq:ppca-rec-term2} into
\eqref{eq:ppca-rec-decomp} and using $\bSigma_{ii}\le\bSigma_{\max}$ on $\mathcal{A}$,
\begin{equation}
\big|\theta_i - \hat{\theta}_i\big|
\le \bSigma_{\max}\sqrt{\frac{2\ln(4d/\delta)}{cn}} + \epsilon(\delta/2).
\label{eq:ppca-rec-branch1}
\end{equation}
Separately, $0 \le \theta_i \le \bSigma_{ii}$ and $0 \le \hat{\theta}_i \le \hat{\bSigma}_{ii}$
by \eqref{eq:PPCA-sig-dicomp}, so on $\mathcal{A}$,
\begin{equation}
\begin{split}
\big|\theta_i - \hat{\theta}_i\big|
&\le \big|(\hat{\bSigma}-\bSigma)_{ii}\big| + 2\bSigma_{ii} \\
&\le \bSigma_{\max}\sqrt{\frac{2\ln(4d/\delta)}{cn}} + 2\bSigma_{\max}.
\end{split}
\label{eq:ppca-rec-branch2}
\end{equation}
Taking the smaller of \eqref{eq:ppca-rec-branch1} and \eqref{eq:ppca-rec-branch2} gives, for every
$i \in \{1,\dots,d\}$ simultaneously on $\mathcal{A}\cap\mathcal{B}$,
\begin{equation}
\big|\theta_i - \hat{\theta}_i\big| \le B(\delta),
\end{equation}
with $B(\delta)$ as in \eqref{eq:ppca-Bmax}. Together with \eqref{eq:ppca-rec-AB},
\begin{equation}
\mathbb{P}(\mathcal{E}) \ge 1-\delta.
\end{equation}

On $\mathcal{E}$, for any $s \in S$ and $j \notin S$,
\begin{equation}
\begin{split}
\hat{\theta}_{s} - \hat{\theta}_{j}
&\ge \big(\theta_{s} - B(\delta)\big) - \big(\theta_{j} + B(\delta)\big) \\
&= \big(\theta_{s} - \theta_{j}\big) - 2B(\delta) \\
&\ge \min_{s' \in S}\theta_{s'} - \max_{j' \notin S}\theta_{j'} - 2B(\delta)
\;>\; 0,
\end{split}
\end{equation}
the last inequality by \eqref{eq:ppca-gap}. Hence
\begin{equation}
\min_{s \in S}\hat{\theta}_{s} > \max_{j \notin S}\hat{\theta}_{j},
\end{equation}
which is \eqref{eq:ppca-recovery-claim}.

For the SNR statement, \eqref{eq:cest} and \eqref{def:SNR} give
\begin{equation}
\bSNR_i = \frac{\theta_i}{\sigma^2},
\qquad
\widehat{\bSNR}_i = \frac{\hat{\theta}_i}{\hat{\sigma}^2},
\qquad
\hat{\sigma}^2 > 0,
\end{equation}
where neither $\sigma^2$ nor $\hat{\sigma}^2$ depends on $i$. Therefore
\begin{align*}
\bSNR_i > \bSNR_j &\iff \theta_i > \theta_j,
\\
\widehat{\bSNR}_i > \widehat{\bSNR}_j &\iff \hat{\theta}_i > \hat{\theta}_j,
\end{align*}
and dividing \eqref{eq:ppca-gap} by $\sigma^2$, together with $\theta_j = 0$ for $j \notin S$,
yields \eqref{eq:ppca-gap-snr}.
\end{proof}

The latent factor models are generally non-identifiable without additional constraints. 
Therefore, authors have introduced additional constraints (\textbf{IC1-IC5}) in \cite{bai2012statistical} to ensure the full identifiability of the model parameters. 
Now we will state the theorem that proves the consistency of the estimators and establishes their rates of convergence in the high high-dimensional data regime:

\begin{theorem}[Theorem $5.1$ from \cite{bai2012statistical}]\label{thm:lfa-bai}
Under Assumptions (\textbf{A3-A6}), when $d,n \to \infty$, with any one of the identification conditions (\textbf{IC1-IC5}) mentioned in \cite{bai2012statistical}, we have:
\begin{align}
\frac{1}{d}\sum_{i =1 }^d\frac{1}{\hat{\sigma}^2_i}\|\hat{\bW}_{i,\cdot} - \bW_{i,\cdot}\|^2&= O_p(\frac{1}{n}),\\
\frac{1}{d}\sum_{i = }^d(\hat{\sigma}_i^2 - \sigma_i^2)^2&= O_p(\frac{1}{n}),
\end{align}
where, $X_n = O_p\left(\frac{1}{n}\right)$ means for any $\epsilon > 0$, there exists a finite, positive constant $M$ such that for all $n \ge 1$: $P\left( \left| nX_n \right| > M \right) < \epsilon$.
\end{theorem}
Now, we will prove the following lemma, which is a direct implication of Theorem \ref{thm:lfa-bai}. However, the constraints (\textbf{IC1-IC5}) to ensure identifiability have only been applied to prove asymptotic results for $|\hat{\bW}-\bW|$, not for $|\hat{\sigma}_i^2 - \sigma_i^2|$; therefore, the constraints (\textbf{IC1-IC5}) have also not been considered in Lemma \ref{lem:lfa-noise1}.
\begin{lem}[\textbf{Asymptotic Bound for Noise Varaiance Estimation Error}]\label{lem:lfa-noise1}
    Under the assumptions \textbf{A3-A6}, as $(d,n) \to \infty$, $ \max_{i=1,\dots,d} |\hat{\sigma}_i^2 - \sigma_i^2| = O_P\left( \sqrt{\frac{d}{n}} \right)$.
\end{lem}
\begin{proof}
    Let $\hat{\bPsi} = \mathrm{diag}(\hat{\sigma}_1^2, \dots, \hat{\sigma}_d^2)$ be the estimated noise variance matrix. From Theorem \ref{thm:lfa-bai},
\begin{equation} \label{eq:bai-variance}
    \frac{1}{d} \sum_{i=1}^d (\hat{\sigma}_i^2 - \sigma_i^2)^2 = O_P\left(\frac{1}{n}\right).
\end{equation}

From this, we can derive the Frobenius norm bound for the difference between $\hat{\bPsi}$ and $\bPsi$:
\begin{align*}
\| \hat{\bPsi} - \bPsi \|_F^2 &= \sum_{i=1}^d (\hat{\sigma}_i^2 - \sigma_i^2)^2 = d \cdot \left( \frac{1}{d} \sum_{i=1}^d (\hat{\sigma}_i^2 - \sigma_i^2)^2 \right) \\
&= d \cdot O_P\left( \frac{1}{n} \right) = O_P\left( \frac{d}{n} \right).
\end{align*}
Taking the square root, we get:
\begin{equation}
    \| \hat{\bPsi} - \bPsi \|_F = O_P\left( \sqrt{\frac{d}{n}} \right).
\end{equation}

From the Frobenius norm bound, we can directly derive a bound for the maximum element-wise deviation. 
Since the Frobenius norm is the square root of the sum of squared elements, the largest squared element must be less than or equal to the sum of all squared elements:
\[ \max_{i=1,\dots,d} (\hat{\sigma}_i^2 - \sigma_i^2)^2 \leq \sum_{i=1}^d (\hat{\sigma}_i^2 - \sigma_i^2)^2 = \| \hat{\bPsi} - \bPsi \|_F^2. \]
Substituting the Frobenius norm bound:
\[ \max_{i=1,\dots,d} (\hat{\sigma}_i^2 - \sigma_i^2)^2 = O_P\left( \frac{d}{n} \right). \]
Taking the square root, we get the bound for the maximum absolute deviation:
\begin{equation}
    \max_{i=1,\dots,d} |\hat{\sigma}_i^2 - \sigma_i^2| = O_P\left( \sqrt{\frac{d}{n}} \right).
\end{equation}
\end{proof}




\begin{proof} of Theorem \ref{lem:lfa-fixedpoint}. 
The M-step (of the EM algorithm) update equation for $\mathbf{W}$ is given in Theorem \ref{thm:lfa}, with converged $\mathbf{W}$):
\[ \hat{\mathbf{W}} = \left(\sum_{i=1}^{n}\mathbf{\bx}_{i}E(\bgamma_i|\mathbf{x}_{i})^{T}\right)\left(\sum_{i=1}^{n}E(\bgamma_i\bgamma_i^{T}|\mathbf{x}_{i})\right)^{-1} \]

Let's simplify the terms:
\begin{itemize}
    \item \textbf{Numerator sum:}
    Substituting $E(\bgamma_i|\mathbf{x}_{i}) = \bbeta \mathbf{x}_{i}$:
    \begin{align*} \sum_{i=1}^{n}\mathbf{x}_{i}E(\bgamma_i|\mathbf{x}_{i})^{T} &= \sum_{i=1}^{n}\mathbf{x}_{i}(\bbeta \mathbf{x}_{i})^{T} =\sum_{i=1}^{n}\mathbf{x}_{i}\mathbf{x}_{i}^{T}\bbeta^{T} \\ &= \left(\sum_{i=1}^{n}\mathbf{x}_{i}\mathbf{x}_{i}^{T}\right)\bbeta^{T} = (n\mathbf{\hat{\Sigma}})\bbeta^{T}.\end{align*}

    \item \textbf{Denominator sum:}
    Substituting $E(\bgamma_i\bgamma_i^{T}|\mathbf{x}_{i}) = \mathbf{I}_{r}-\bbeta \hat{\mathbf{W}}+\bbeta \mathbf{x}_{i}\mathbf{x}_{i}^{T}\bbeta^{T}$:
    \begin{align*} \sum_{i=1}^{n}E(\bgamma_i\bgamma_i^{T}|\mathbf{x}_{i}) &= \sum_{i=1}^{n}(\mathbf{I}_{r}-\bbeta \hat{\mathbf{W}}+\bbeta \mathbf{x}_{i}\mathbf{x}_{i}^{T}\bbeta^{T}) \\ &= n(\mathbf{I}_{r}-\bbeta \hat{\mathbf{W}}) + \bbeta\left(\sum_{i=1}^{n}\mathbf{x}_{i}\mathbf{x}_{i}^{T}\right)\bbeta^{T} \\ &= n(\mathbf{I}_{r}-\bbeta \hat{\mathbf{W}}) + \bbeta(n\mathbf{\hat{\Sigma}})\beta^{T} \\ &= n(\mathbf{I}_{r}-\bbeta \hat{\mathbf{W}} + \bbeta \mathbf{\hat{\Sigma}} \beta^{T}).
    \end{align*}
\end{itemize}

Now, substitute these simplified sums back into the $\hat{\mathbf{W}}$ equation:
\begin{align*} \hat{\mathbf{W}} &= (n\mathbf{\hat{\Sigma}})\bbeta^{T} (n(\mathbf{I}_{r}-\bbeta \hat{\mathbf{W}} + \bbeta \mathbf{\hat{\Sigma}} \beta^{T}))^{-1} \\ &= (n\mathbf{\hat{\Sigma}})\bbeta^{T} \frac{1}{n}(\mathbf{I}_{r}-\bbeta \hat{\mathbf{W}} + \bbeta \mathbf{\hat{\Sigma}} \bbeta^{T})^{-1} \\ &= \mathbf{\hat{\Sigma}}\bbeta^{T} (\mathbf{I}_{r}-\bbeta \hat{\mathbf{W}} + \bbeta \mathbf{\hat{\Sigma}} \bbeta^{T})^{-1}.
\end{align*}
Multiplying both sides by $(\mathbf{I}_{r}-\beta \hat{\mathbf{W}} + \bbeta \mathbf{\hat{\Sigma}} \bbeta^{T})$ from the right, we obtain:
\[ \hat{\mathbf{W}}(\mathbf{I}_{r}-\bbeta \hat{\mathbf{W}} + \beta \mathbf{\hat{\Sigma}} \bbeta^{T}) = \mathbf{\hat{\Sigma}}\bbeta^{T}.
\]

To further apply matrix algebra, we substitute $\bbeta = \hat{\mathbf{W}}^{T}(\hat{\mathbf{\Psi}}+\hat{\mathbf{W}}\hat{\mathbf{W}}^{T})^{-1}$. Let $\mathbf{M} = \hat{\mathbf{\Psi}}+\hat{\mathbf{W}}\hat{\mathbf{W}}^{T}$, so $\bbeta = \hat{\mathbf{W}}^{T}\mathbf{M}^{-1}$. 
Also note that $\hat{\mathbf{W}}\hat{\mathbf{W}}^{T} = \mathbf{M} - \hat{\mathbf{\Psi}}$.

The identity becomes:

\begin{multline} 
\hat{\mathbf{W}} - \hat{\mathbf{W}}\bbeta \hat{\mathbf{W}} + \hat{\mathbf{W}}\bbeta \mathbf{\hat{\Sigma}} \bbeta^{T} = \mathbf{\hat{\Sigma}}\bbeta^{T}, \\ 
\hat{\mathbf{W}} - \hat{\mathbf{W}}(\hat{\mathbf{W}}^{T}\mathbf{M}^{-1}) \hat{\mathbf{W}} + \hat{\mathbf{W}}(\hat{\mathbf{W}}^{T}\mathbf{M}^{-1}) \mathbf{\hat{\Sigma}} (\mathbf{M}^{-1}\hat{\mathbf{W}}) = \mathbf{\hat{\Sigma}}\mathbf{M}^{-1}\hat{\mathbf{W}}, \\ \hat{\mathbf{W}} - (\hat{\mathbf{W}}\hat{\mathbf{W}}^{T})\mathbf{M}^{-1} \hat{\mathbf{W}} + (\hat{\mathbf{W}}\hat{\mathbf{W}}^{T})\mathbf{M}^{-1} \mathbf{\hat{\Sigma}} \mathbf{M}^{-1} \hat{\mathbf{W}} = \mathbf{\hat{\Sigma}}\mathbf{M}^{-1}\hat{\mathbf{W}}, \\ \hat{\mathbf{W}} - (\mathbf{M} - \hat{\mathbf{\Psi}})\mathbf{M}^{-1} \hat{\mathbf{W}} + (\mathbf{M} - \hat{\mathbf{\Psi}})\mathbf{M}^{-1} \mathbf{\hat{\Sigma}} \mathbf{M}^{-1} \hat{\mathbf{W}} = \mathbf{\hat{\Sigma}}\mathbf{M}^{-1}\hat{\mathbf{W}}, \\ \hat{\mathbf{W}} - (I - \hat{\mathbf{\Psi}}\mathbf{M}^{-1}) \hat{\mathbf{W}} + (I - \hat{\mathbf{\Psi}}\mathbf{M}^{-1}) \mathbf{\hat{\Sigma}} \mathbf{M}^{-1} \hat{\mathbf{W}} = \mathbf{\hat{\Sigma}}\mathbf{M}^{-1}\hat{\mathbf{W}}.
\end{multline}

Simplifying, moving all terms to one side and factoring out $\mathbf{M}^{-1}\hat{\mathbf{W}}$, we obtain:
\[ 
(\hat{\mathbf{\Psi}} + \mathbf{\hat{\Sigma}} - \hat{\mathbf{\Psi}}\mathbf{M}^{-1}\mathbf{\hat{\Sigma}} - \mathbf{\hat{\Sigma}})\mathbf{M}^{-1}\hat{\mathbf{W}} = \bzero.
\]
This simplifies to:
\[ 
(\hat{\mathbf{\Psi}} - \hat{\mathbf{\Psi}}\mathbf{M}^{-1}\mathbf{\hat{\Sigma}})\mathbf{M}^{-1}\hat{\mathbf{W}} = \bzero, 
\]
\[ 
\hat{\mathbf{\Psi}}(\mathbf{I} - \mathbf{M}^{-1}\mathbf{\hat{\Sigma}})\mathbf{M}^{-1}\hat{\mathbf{W}} = \bzero.
\]

Assuming $\hat{\mathbf{\Psi}}$ (\textbf{A9(ii),A6}) is invertible, we get
\begin{equation}
\boxed{(\mathbf{I} - \mathbf{M}^{-1}\mathbf{\hat{\Sigma}})\mathbf{M}^{-1}\hat{\mathbf{W}} = \bzero}.   \label{eq:W}
\end{equation} 
This identity must hold at the maximum likelihood estimate for $\hat{\mathbf{W}}$ and $\hat{\mathbf{\Psi}}$.
\\
We next focus on the identity for $\hat{\mathbf{\Psi}}$.
The M-step update equation for $\hat{\mathbf{\Psi}}$ is given as in Theorem \ref{thm:lfa}, with converged $\hat{\mathbf{\Psi}}$:
\begin{equation}\label{eq:lfa-noise-var-em}
    \hat{\mathbf{\Psi}} = \frac{1}{n}D\left(\sum_{i=1}^{n}\mathbf{x}_{i}\mathbf{x}_{i}^{T}-\hat{\mathbf{W}}E(\bgamma_i|\mathbf{x}_{i})\mathbf{x}_{i}^{T}\right)
\end{equation}
Let's simplify the sum term inside $D(\cdot)$:
\begin{align*} \sum_{i=1}^{n}\mathbf{x}_{i}\mathbf{x}_{i}^{T}-\hat{\mathbf{W}}E(\bgamma_i|\mathbf{x}_{i})\mathbf{x}_{i}^{T} &= \sum_{i=1}^{n}\mathbf{x}_{i}\mathbf{x}_{i}^{T} - \hat{\mathbf{W}}\sum_{i=1}^{n}E(\bgamma_i|\mathbf{x}_{i})\mathbf{x}_{i}^{T} \\ &= n\mathbf{\hat{\Sigma}} - \hat{\mathbf{W}}\sum_{i=1}^{n}(\bbeta \mathbf{x}_{i})\mathbf{x}_{i}^{T} \\ &= n\mathbf{\hat{\Sigma}} - \hat{\mathbf{W}}\bbeta\left(\sum_{i=1}^{n}\mathbf{x}_{i}\mathbf{x}_{i}^{T}\right) \\ &= n\mathbf{\hat{\Sigma}} - \hat{\mathbf{W}}\bbeta(n\mathbf{\hat{\Sigma}}).
\end{align*}
Substitute this back into the $\hat{\mathbf{\Psi}}$ equation:
\begin{align*} 
\hat{\mathbf{\Psi}} &= \frac{1}{n}D(n\mathbf{\hat{\Sigma}} - \hat{\mathbf{W}}\bbeta(n\mathbf{\hat{\Sigma}})) \\ &= D(\mathbf{\hat{\Sigma}} - \hat{\mathbf{W}}\bbeta\mathbf{\hat{\Sigma}}).
\end{align*}

Now, substitute $\bbeta = \hat{\mathbf{W}}^{T}\mathbf{M}^{-1}$:
\[ \hat{\mathbf{\Psi}} = D(\mathbf{\hat{\Sigma}} - \hat{\mathbf{W}}(\hat{\mathbf{W}}^{T}\mathbf{M}^{-1})\mathbf{\hat{\Sigma}}), \]
\[ \hat{\mathbf{\Psi}} = D(\mathbf{\hat{\Sigma}} - \hat{\mathbf{W}}\hat{\mathbf{W}}^{T}\mathbf{M}^{-1}\mathbf{\hat{\Sigma}}), 
\]
Substitute $\hat{\mathbf{W}}\hat{\mathbf{W}}^{T} = \mathbf{M} - \hat{\mathbf{\Psi}}$:
\begin{align*} \hat{\mathbf{\Psi}} &= D(\mathbf{\hat{\Sigma}} - (\mathbf{M} - \hat{\mathbf{\Psi}})\mathbf{M}^{-1}\mathbf{\hat{\Sigma}}) \\ &= D(\mathbf{\hat{\Sigma}} - (\mathbf{I} - \hat{\mathbf{\Psi}}\mathbf{M}^{-1})\mathbf{\hat{\Sigma}}) \\ &= D(\mathbf{\hat{\Sigma}} - \mathbf{\hat{\Sigma}} + \hat{\mathbf{\Psi}}\mathbf{M}^{-1}\mathbf{\hat{\Sigma}}).
\end{align*}
The $\mathbf{\hat{\Sigma}}$ terms cancel out:
\[ \hat{\mathbf{\Psi}} = D(\hat{\mathbf{\Psi}}\mathbf{M}^{-1}\mathbf{\hat{\Sigma}}). 
\]
This is a critical identity. Given that $\hat{\mathbf{\Psi}}$ is a diagonal matrix, we can state that its $i$-th diagonal element  is equal to the $i$-th diagonal element of $\hat{\mathbf{\Psi}}\mathbf{M}^{-1}\mathbf{\hat{\Sigma}}$.

If $\hat{\mathbf{\Psi}}$ is invertible (\textbf{A6}), we can imply a further identity by "dividing" by $\hat{\mathbf{\Psi}}$ (element-wise on the diagonal operation):
\begin{equation}\label{CS}
    \boxed{\mathbf{I} = D(\mathbf{M}^{-1}\mathbf{\hat{\Sigma}})}.
\end{equation}

This identity states that the diagonal elements of the product $\mathbf{M}^{-1}\mathbf{\hat{\Sigma}}$ must be equal to 1. Since $\mathbf{M}^{-1}\mathbf{\hat{\Sigma}}$ is symmetric, this is equivalent to saying that $(\mathbf{M}^{-1}\mathbf{\hat{\Sigma}})_{ii} = 1$ for all $i=1, \dots, d$.

Now we will proceed to prove the next part of the theorem.
Let $\bA=\mathbf{\hat{\Sigma}}-\bM$.
 Then $\mathbf{\hat{\Sigma}}=\bM + \bA$ and $\bM^{-1}\mathbf{\hat{\Sigma}}=\mathbf{I+M^{-1}A}$ so Eq. \eqref{eq:W} becomes
\[
\bM^{-1}\bA\bM^{-1}\bW=0, 
\]
which after multiplying to the left with $\bM$ becomes
\[
\bA\bM^{-1}\bW=0.
\]
Multiply by $\bW^T$ to the right and we obtain
\[
\bA\bM^{-1}\bW\bW^T=\bA\bM^{-1}(\bM-\bPsi)=0.
\]
which means
\[
\bA=\bA\bM^{-1}\bPsi.
\]
The identity from \eqref{CS} becomes
\[
(\bM^{-1}\bA)_{ii}=0.
\]
But $\bA=\bA^T$ so $(\bA\bM^{-1})_{ii}=0$. But in this case, since $\bA=\bA\bM^{-1}\bPsi$, we also have that $\bA_{ii}=0$ for all $i$.

This implies that:
\begin{equation}
    \boxed{(\hat{\mathbf{W}}\hat{\mathbf{W}}^{T})_{ii}=\mathbf{\hat{\Sigma}}_{ii}-\hat\bPsi_{ii}.}
\end{equation}

Let us proceed to the last part of the theorem.
\begin{equation}
\begin{split}
\abs{ (\bW\bW^T)_{ii} - (\hat{\bW}\hat{\bW}^T)_{ii} } &= \abs{ (\mathbf{\Sigma_{ii}} - \bPsi_{ii}) - (\mathbf{\hat{\Sigma}}_{ii} - \hat{\bPsi}_{ii}) } \\
&= \abs{(\mathbf{\mathbf{\Sigma_{ii}} - \mathbf{\hat{\Sigma}}_{ii}) + (\hat{\bPsi}_{ii} - \bPsi_{ii}) }} \\
&\le \abs{(\mathbf{\hat{\Sigma}}- \bSigma)_{ii}} + \abs{ \hat{\bPsi}_{ii} - \bPsi_{ii} }.
\label{eq:lfa_error_decomp_2}    
 \end{split}
\end{equation}
\end{proof}

\begin{proof} of Theorem \ref{thm:asym}.
From Lemma \ref{lem:lfa-fixedpoint}, we get:
$\abs{ (\bW\bW^T)_{ii} - (\hat{\bW}\hat{\bW}^T)_{ii} } \le \abs{(\mathbf{\hat{\Sigma}}- \bSigma)_{ii}} + \abs{ \hat{\bPsi}_{ii} - \bPsi_{ii}}$.

From Theorem \ref{thm:scalar_bernstein}, we get $\abs{(\mathbf{\hat{\Sigma}}- \bSigma)_{ii}} = O_p(1/\sqrt{n})$ and from Lemma \ref{lem:lfa-noise1}, $\abs{ \hat{\bPsi}_{ii} - \bPsi_{ii}} = O_p(\sqrt{d/n})$.

Therefore, \\
    $\abs{ (\bW\bW^T)_{ii} - (\hat{\bW}\hat{\bW}^T)_{ii} } = O_p(\max(1/\sqrt{n}, \sqrt{d/n})) = O_p(\sqrt{d/n})$

    Now we, turn to SNRs. Let $X_0, Y_0$ be true values and $X, Y$ be estimators. If $1/c \le Y \le c$, for some $c >0$ then,

\begin{align*}
    &\abs{ (X Y_0 - X_0 Y) / (Y Y_0) } \\
    &= \abs{ (X - X_0)Y_0 - X_0(Y - Y_0) } / (|Y| Y_0) \\
    &\le \frac{c}{Y_0} \left( Y_0 |X - X_0| + X_0 |Y - Y_0| \right).
\end{align*}

    We wish to bound the error $\abs{\widehat{\bSNR}_i - \bSNR_i}$. Note that, $\bSNR_i = \frac{\bW\bW^T_{ii}}{\sigma_i^2} = \frac{\bSigma_{ii}-\sigma^2_i}{\sigma_i^2} = \frac{\bSigma_{ii}}{\sigma^2_i}-1$. Therefore, the numerator and denominator terms are:
    \begin{itemize}
    \item True Numerator: $X_0 = \bSigma_{ii}$
    \item Estimated Numerator: $X = \hat{\bSigma}_{ii}$
    \item True Denominator: $Y_0 = \sigma_i^2$
    \item Estimated Denominator: $Y = \hat{\sigma}_i^2$
    \end{itemize}
    Assumption \textbf{A6} assures that there is a $c$, such that $1/c \le \sigma_i^2 \le c$ for sufficiently large $n$. Therefore, we get the following:
    \begin{align*}
\abs{\widehat{\bSNR}_i - \bSNR_i} &= \abs{\frac{X}{Y} - \frac{X_0}{Y_0}} \notag \\
&\le \frac{c}{Y_0} \left( Y_0 \abs{X - X_0} + X_0 \abs{Y - Y_0} \right) \notag \\
&= c(\abs{\hat{\bSigma}_{ii} - \bSigma_{ii}} + \bSNR_i\abs{\hat{\sigma}_i^2 - \sigma_i^2})\\
&=c(O_p(\sqrt{1/n}) + \bSNR_i \times O_p(\sqrt{d/n})) \\
&= O_p(\sqrt{d/n}).
\end{align*}
\end{proof}

\subsection{Discussion on the large $d$, large $n$ case}

In this type of results, both the number of features $d$ and the number of samples $n$ tend to infinity, while the ratio $\frac{d}{n}( = \gamma \ge 0)$ is kept constant.  
This is also known as the “ultra-high dimensional” or “big data” regime. 
The main challenge to drawing inferences on asymptotic behaviors of eigenvalues and eigenvectors in this setup is that the sample covariance matrix does not well approximate the population covariance matrix unlike the case when $d$ was fixed. 
There has been considerable effort to establish convergence results for sample eigenvalues and eigenvectors in recent years. 
Some of these findings will be discussed below.

Recent theoretical advancements on the 'large $d$ large $n$' setup are based on the assumption that the data matrix, $\textbf{X}$ is generated from a spiked population covariance model. 
The notion of the spiked population covariance model was first introduced by \cite{johnstone2001distribution}.

\begin{definition} \label{spiked}
\textbf{Spiked Covariance model.} Under this model, the data matrix $\textbf{X}$, can be viewed as $\boldsymbol{X}^T = \boldsymbol{E\Lambda}^{\frac{1}{2}}\boldsymbol{Z}$, where $\boldsymbol{E} = [\boldsymbol{e_1},\boldsymbol{e_2},\cdots,\boldsymbol{e_d}]$ is a $d \times d$ orthogonal matrix, $\boldsymbol{\Lambda} = \diag(\lambda_1,\lambda_2,\cdots,\lambda_d)$ with $\lambda_1 \geq \lambda_2 \geq \cdots \geq \lambda_d$ and \textbf{Z} is a $d \times n$ matrix constructed with iid random variables $\boldsymbol{Z}_{ij}$ with $E(\boldsymbol{Z}_{ij}) = 0$, $E(\boldsymbol{Z}_{ij}^2) = 1$ and $E(\boldsymbol{Z}_{ij}^4) \le \infty$. The population covariance matrix is $\boldsymbol{\Sigma} = \boldsymbol{E\Lambda E^T}$. Here, $\lambda_k$'s are assumed to follow a specific structure, $\lambda_1 \ge \lambda_2 \ge \cdots  \ge \lambda_m > \lambda_{m+1} = \cdots = \lambda_d = 1$.
\end{definition}
 
The sample covariance matrix is $\hat{\bSigma} = \boldsymbol{X^T X}/n = \boldsymbol{E\Lambda}^{\frac{1}{2}}\boldsymbol{Z} \boldsymbol{Z}^T \boldsymbol{\Lambda^{\frac{1}{2}}E}^T/n$.

The spectral decomposition of the sample covariance matrix $\hat{\bSigma}$ is, 
$\hat{\bSigma} = \boldsymbol{USU^T}$. Here, $\boldsymbol{S} = \diag(s_1,s_2,\cdots s_d)$ are the ordered sample eigenvalues and $\boldsymbol{U} = [\boldsymbol{u_1}, \boldsymbol{u_2}, \cdots, \boldsymbol{u_d}]$ is the corresponding $d \times d$ sample eigenvector matrix. 
For the remaining of the section, we assume that $\lim_{n \to \infty}\frac{d}{n} = \gamma$. Also there are $k$ population eigenvalues such that $\lambda_i > 1+\sqrt{\gamma}$, for $i \le k$

The following result is due to \cite{baik2006eigenvalues}.

\begin{theorem}\label{Theo1}
[from \cite{baik2006eigenvalues}] For $\gamma \in (0,1)$, the following holds:
\[
s_i \overset{a.s.}{\to}
    \begin{cases}
        \rho(\lambda_i), & \text{if } i\le k\\
        (1+\sqrt{\gamma})^2 & \text{otherwise},\\
    \end{cases} 
\]
where $\rho(x) = x(1 + \frac{\gamma}{x-1})$.
\end{theorem}

It is evident from Theorem \ref{Theo1} that the sample eigenvalues are not consistent estimates of the population counterparts. 
However, a consistent estimator can be found for $\lambda_i > 1+\gamma$ using the following inverse function:
\begin{equation}
    \rho^{-1}(x) = \frac{x+1-\gamma + \sqrt{(x+1-\gamma)^2 - 4x}}{2}
\end{equation}
Also, it has been shown in \cite{bai2008central} that $s_i$ are asymptotically normal. 

Although consistency could not be proved for $\gamma > 0$, \cite{lee2010convergence} proved consistency for $\gamma = 0$.

\begin{lemma}
[due to \cite{lee2010convergence}]\label{lemma:0}
If $\lim_{n \to \infty}\frac{d}{n} = \gamma = 0$, then,
\[
s_i \overset{a.s.}{\to} 
    \begin{cases}
        \lambda_i & \text{if } i\le m\\
        1 & \text{otherwise.}\\
    \end{cases} 
\]
\end{lemma}

For eigenvectors, the convergence of the angle between sample eigen vectors $\bu_i$ and population eigenvectors $\be_i$ has been proved for Gaussian $\boldsymbol{Z}_{ij}$'s in \cite{paul2007asymptotics}. 
The author used the inner product between two unit vectors to represent the cosine angle between $\bu_i$ and $\be_i$. 

\begin{theorem}\label{angle}
[due to \cite{paul2007asymptotics}] 
Under the assumption, that multiplicity of eigen values are one, if $\lim_{n \to \infty} \frac{d}{n} = \gamma \in (0,1)$, and $\boldsymbol{Z_{ij}}$'s follow the standard normal distribution, then

\[
|<\boldsymbol{e}_i, \boldsymbol{u}_i>| \overset{a.s.}{\to} 
    \begin{cases}
        \phi(\lambda_i) & \text{if } \lambda_i > 1+\sqrt{\gamma}\\
        0 & \text{if } 1 < \lambda_i < 1+\sqrt{\gamma}.
    \end{cases} 
\]
\end{theorem}

Here, $<\boldsymbol{a},\boldsymbol{b}>$ represents the inner product between two vectors $\boldsymbol{a}$ and $\boldsymbol{b}$, and $\phi(x) = \sqrt{\frac{1-\frac{\gamma}{(x-1)^2}}{1+\frac{\gamma}{x-1}}}$.

The same conclusion was reached in \cite{nadler2008finite} for $\gamma > 0$ using a matrix perturbation approach under the Gaussian random noise model. 
Theorem \ref{angle} was generalized in \cite{lee2010convergence} by relaxing the distributional assumption and proving a weaker convergence (in probability) for the angles between population and sample eigenvectors, when $\gamma \ge 0$.

\begin{lemma}
[due to \cite{lee2010convergence}]Under the assumption of multiplicity one, if $\lim_{n \to \infty} \frac{d}{n} = \gamma \ge 0$, 
\[
|<\boldsymbol{e}_i, \boldsymbol{u}_i>| \overset{p}{\to} 
    \begin{cases}
        \phi(\lambda_i) & \text{if } \lambda_i > 1+\sqrt{\gamma}\\
        0 & \text{if } 1 < \lambda_i < 1+\sqrt{\gamma}.
    \end{cases} 
\]
\end{lemma}

\end{document}